\documentclass[11pt,a4paper,logo]{googledeepmind}

\setleftlogo[120pt]{imgs/logo-removebg-preview-Photoroom.jpg} 
\setrightlogo[180pt]{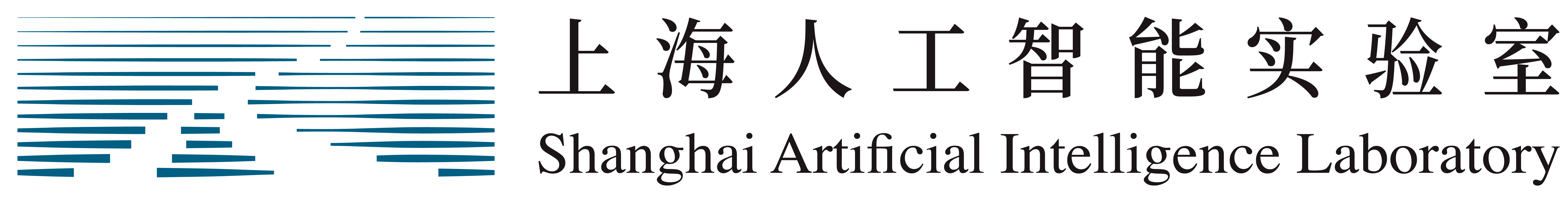}

\usepackage[T1]{fontenc}
\usepackage{pifont}
\usepackage[
    natbib=true,
    backend=biber,
    style=numeric, % Changed from 'alphabetic' to 'numeric' as per common technical report styles
    sorting=none % To keep references in order of appearance or as defined in .bib
]{biblatex}
\AtEveryBibitem{\clearfield{month}}
\AtEveryBibitem{\clearfield{day}}

\usepackage{csquotes}

\newcommand{\ProjectName}{FlowSearch}

\title{\ProjectName: Advancing Deep Research with Dynamic Structured Knowledge Flow}

\correspondingauthor{
$\spadesuit$: Core Contributor (listed in random order) \quad
$\clubsuit$: Corresponding Author \\ 
Please send correspondence regarding this report to zhangbo@pjlab.org.cn
}

\author[1 $\spadesuit$]{Yusong Hu}
\author[1 $\spadesuit$]{Runmin Ma}
\author[1]{Yue Fan}
\author[1]{Jinxin Shi}
\author[1]{Zongsheng Cao}
\author[1]{Yuhao Zhou}
\author[1]{Jiakang Yuan}
\author[1]{Shuaiyu Zhang}
\author[1]{Shiyang Feng}
\author[1]{Xiangchao Yan}
\author[1]{Shufei Zhang}
\author[1]{Wenlong Zhang}
\author[1 $\clubsuit$]{Lei Bai}
\author[1 $\clubsuit$]{Bo Zhang}

\affil[1]{Shanghai Artificial Intelligence Laboratory}

\usepackage{pdflscape}

\usepackage{textcomp}
\usepackage{rotating}

\usepackage{setspace}
\usepackage{microtype} 

\usepackage{soul} 

\usepackage{graphicx}
\usepackage{subcaption}
\usepackage{caption}

\usepackage{booktabs}
\usepackage[table]{xcolor}
\sethlcolor{green!14}

\usepackage{array}
\usepackage{multirow}

\usepackage{amsmath}
\usepackage{siunitx}

\usepackage{enumitem}
\usepackage{float}
\usepackage{seqsplit}
\usepackage{framed}

\usepackage{tikz}
\usepackage{hyperref}
\usepackage{url}
\usepackage{listings}
\usepackage{xcolor}
\usepackage{soul}
\usepackage{colortbl}  
\usepackage{graphicx}
\usepackage{wrapfig}
\usepackage[most]{tcolorbox}
\usepackage{enumitem}
\usepackage{hyperref}

\tcbuselibrary{listingsutf8}
\definecolor{mycolor}{RGB}{50,80,150}

\usepackage{ragged2e}
\usepackage[most]{tcolorbox}

\usepackage{makecell}
\usepackage{adjustbox}

\usepackage[symbol]{footmisc}
\newcolumntype{Y}{>{\RaggedRight\arraybackslash}X}

\newcommand{\pdfdiff}[1]{\textcolor{black}{#1}}

\usepackage{hyperref}
\hypersetup{
    colorlinks=true,
    linkcolor=blue, % Changed for better visibility of links
    citecolor=blue,  % Changed for better visibility of citations
    filecolor=black,
    urlcolor=blue    % Changed for better visibility of URLs
}

\usepackage{tocloft}
\usepackage{etoolbox}
\usepackage{arydshln}
\makeatletter
\patchcmd{\@tocline}
    {\hfil}
    {\leaders\hbox{\hfil}\hfil}
    {}{}
\makeatother

\begin{abstract}
Deep research is an inherently challenging task that demands both breadth and depth of thinking. It involves navigating diverse knowledge spaces and reasoning over complex, multi-step dependencies, which presents substantial challenges for agentic systems. To address this, we propose FlowSearch, a multi-agent framework that actively constructs and evolves a dynamic structured knowledge flow to drive subtask execution and reasoning. FlowSearch is capable of strategically planning and expanding the knowledge flow to enable parallel exploration and hierarchical task decomposition, while also adjusting the knowledge flow in real time based on feedback from intermediate reasoning outcomes and insights.  FlowSearch achieves competitive performance on both general and scientific benchmarks, including GAIA, HLE, GPQA and TRQA, demonstrating its effectiveness in multi-disciplinary research scenarios and its potential to advance scientific discovery. The code is available at \url{https://github.com/InternScience/InternAgent}.

\end{abstract}

\begin{document}
\sloppy
\maketitle

\section{Introduction}
\label{sec:introd}

The general capabilities of Large Language Models (LLMs) enable agent systems for diverse tasks, supporting scientific research and discovery~\citep{schick2023toolformer,fan2024videoagent,team2025novelseek,zhang2025origene}. However, effectively utilizing these capabilities in open-ended research requires iterative hypothesis formulation, strategic information acquisition, and multi-step reasoning in dynamic and uncertain knowledge spaces. These challenges have inspired the development of Deep Research (DR), which combines LLMs with expert-designed knowledge to go beyond basic reasoning or simple information retrieval. Such systems are essential for unlocking the potential of LLMs in enabling scientific discovery on various domains.

Existing deep research systems~\citep{hu2025owl,2025mirothinker} are primarily inspired by either individual or collaborative research paradigms.
\textbf{(1) Single-agent paradigm:}~\citep{wu2025webdancer,tao2025webshaper,li2025webthinker,yao2023react} a single LLM centrally manages the research workflow using a long context window to accumulate and reason over information. While this mirrors individual researchers, it is prone to tunnel vision, overcommitting to early hypotheses and limiting exploratory breadth.
\textbf{(2) Multi-agent paradigm:}~\citep{hu2025owl,manus2025,team2025novelseek} research is scaled via explicit planning and role specialization. However, serial plan execution requires maintaining context across multiple planning steps and agents, forcing aggressive context truncation or retrieval filtering. 
As a result, intermediate reasoning chains are often fragmented, leading to shallow, step-wise reasoning and a loss of global reasoning depth~\citep{huang2506deep}. Overall, both paradigms trade off exploratory breadth against reasoning depth, motivating the need for more adaptive and context-aware research agents.

In this work, we propose a novel Dynamic Structured Knowledge Flow to enable structured and efficient knowledge propagation throughout scientific discovery activities, and instantiate it in FlowSearch, a multi-agent system grounded in this design. Unlike conventional sequential planning, the proposed Structured Knowledge Flow models scientific discovery as \textbf{an evolving graph} (as illustrated in Fig.~\ref{fig:case_study}) of interdependent knowledge units, where exploratory breadth and reasoning depth are jointly supported through incremental expansion and structural revision.
Within this flow, knowledge is decomposed, integrated, and summarized at different granularity levels, enabling deep local reasoning while preserving global coherence. 

FlowSearch instantiates this abstraction as a multi-agent system by operationalizing the dynamic structured knowledge flow throughout the research process. It starts with a flow planner that initializes a graph-structured knowledge flow, where nodes represent subproblems or key concepts and edges capture their knowledge dependencies. As scientific discovery progresses, the knowledge flow is continuously expanded and dynamically revised in response to intermediate findings. Each node orchestrates the execution of corresponding subtasks while supporting recursive decomposition, integration of upstream knowledge, and local summarization, thereby enabling structured and efficient knowledge propagation with effective context management across complex, multi-step scientific problem solving.

\begin{figure}[t]
\begin{center}
\includegraphics[width=\textwidth]{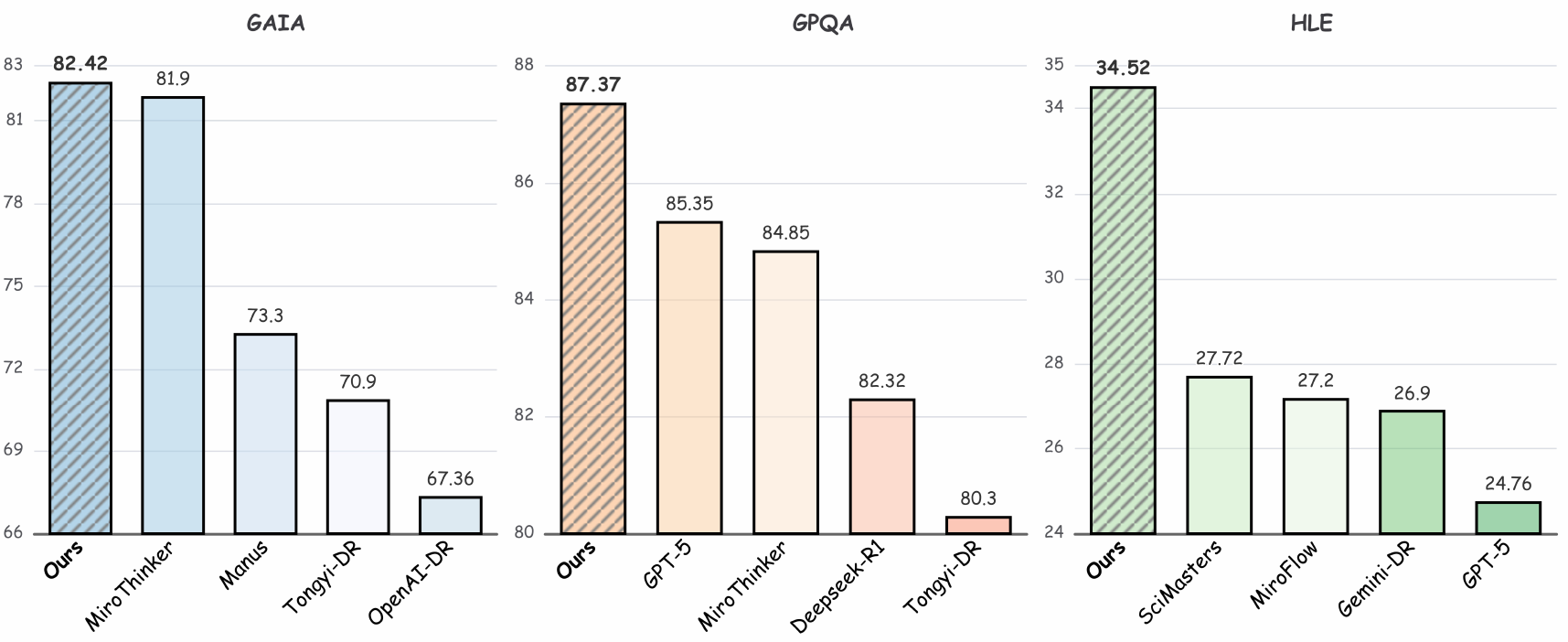} 
\end{center}
\vspace{-4pt}
\caption{FlowSearch (Ours) achieves leading performance on the GAIA, GPQA, and HLE benchmarks, outperforming competitive agentic frameworks (OpenAI-DR, MiroFlow, Tongyi-DR and Manus) as well as LLM-based approaches (GPT-5, DeepSeek-R1, MiroThinker).}
\label{img:comp_introd}
\end{figure}

We conduct experiments on several challenging benchmarks, including GAIA~\citep{mialon2023gaia}, which evaluates the general problem-solving abilities of AI assistants, as well as three scientific-question-answering benchmarks HLE~\citep{phan2025humanity}, GPQA~\citep{rein2024gpqa} and TRQA~\citep{zhang2025origene}. As shown in Fig.~\ref{img:comp_introd}, our approach achieves state-of-the-art results on GAIA, HLE, TRQA and GPQA. These findings highlight the strong problem-solving capability of FlowSearch, enabled by the integration of graph-driven planning. In summary, our main contribution can be described as follows:
\begin{itemize}
    \item Unlike conventional sequential frameworks in deep research agents, we introduce a novel dynamic structured knowledge flow to achieve a trade-off between exploratory breadth and reasoning depth.
    \item We develop FlowSearch, a multi-agent system built upon the dynamic structured knowledge flow, capable of generating structured plans and dynamically refining them during execution phase to enhance performance.
    \item We evaluate FlowSearch on the general AI assistant benchmark GAIA and the multi-disciplinary scientific benchmarks HLE, GPQA and TRQA, demonstrating state-of-the-art results.
\end{itemize}
\section{Related Work}
\subsection{Agentic Systems}

Agentic systems with LLM have evolved from static prompting to perception–action loops, enabling systems to plan~\citep{wang2023plan}, act~\citep{yao2023react}, and learn using external tools. Foundational approaches, such as interleaved reasoning–acting frameworks~\citep{yao2023react} and tree search planning~\citep{yao2023tree}, improve reliability in multi-step tasks, while reflective self-revision mechanisms~\citep{shinn2023reflexion} and external memory~\citep{wang2023voyager} enhance long-horizon consistency. Recent efforts like OpenHands~\citep{wang2025openhands} further expands action spaces and mitigate hallucinations, and evaluates on more realistic interactive benchmarks including AgentBoard~\citep{ma2024agentboard}, StuLife~\citep{cai2025building}, and SWE-bench Verified~\citep{swebench2024verified}. Multi-agent orchestration involves role-specialized collaboration and negotiation~\citep{zhuge2024gptswarm, qiu2025alita}, replacing single-agent end-to-end optimization with a modular and scalable approach.

Despite these advances, most general-purpose agents target short to medium horizon tasks and interactive environments. Scientific research~\citep{openai2025deepresearch}, however, requires handling long-horizon workflows, integrating diverse knowledge, and adapting strategies dynamically. This motivates the development of research-oriented agents, which focus on structured, adaptive, and knowledge-driven scientific inquiry.

\subsection{Deep Research Agents}
Recent advances in DR agents extend LLMs from retrieval-augmented generation to dynamic, tool-driven research workflows. Early systems such as WebGPT~\citep{nakano2021webgpt} and Toolformer~\citep{schick2023toolformer} explored web and API integration, while industrial solutions \textit{e.g.}, OpenAI DR~\citep{openai2025deepresearch}, Gemini DR~\citep{google2024geminidr}, Grok DR~\citep{xai2025grokdeepsearch}, Perplexity DR~\citep{perplexity2025deepresearch}, incorporate adaptive planning, iterative retrieval, and multimodal reasoning. Recently, single-agent designs (\textit{e.g.}, Search-o1~\citep{li2025search}, WebDancer~\citep{wu2025webdancer}, Tongyi DeepResearcher~\citep{qiao2025webresearcher}) enable end-to-end optimization, while multi-agent architectures (\textit{e.g.}, AI Scientist~\citep{lu2024ai}, Agent Laboratory~\citep{schmidgall2025agent}, and InternAgent~\citep{team2025novelseek}) offer modularity and scalability—crucial for complex research.

Recent studies, \textit{e.g.}, GeAR \citep{shen2024gear}, PANGU DeepDiver \citep{shi2025pangudeepdiver}, Alita \citep{qiu2025alita}, also show the benefit of explicit structures and self-evolving mechanisms for multi-hop reasoning. However, the existing DR agents still suffer from sequential bottlenecks and limited hierarchical decomposition, motivating frameworks like our FlowSearch that integrate multi-agent coordination with dynamic structured knowledge flow.

\section{{\ProjectName}}

\begin{figure}[t]
\vspace{-6pt}
\begin{center}
\includegraphics[width=\textwidth]{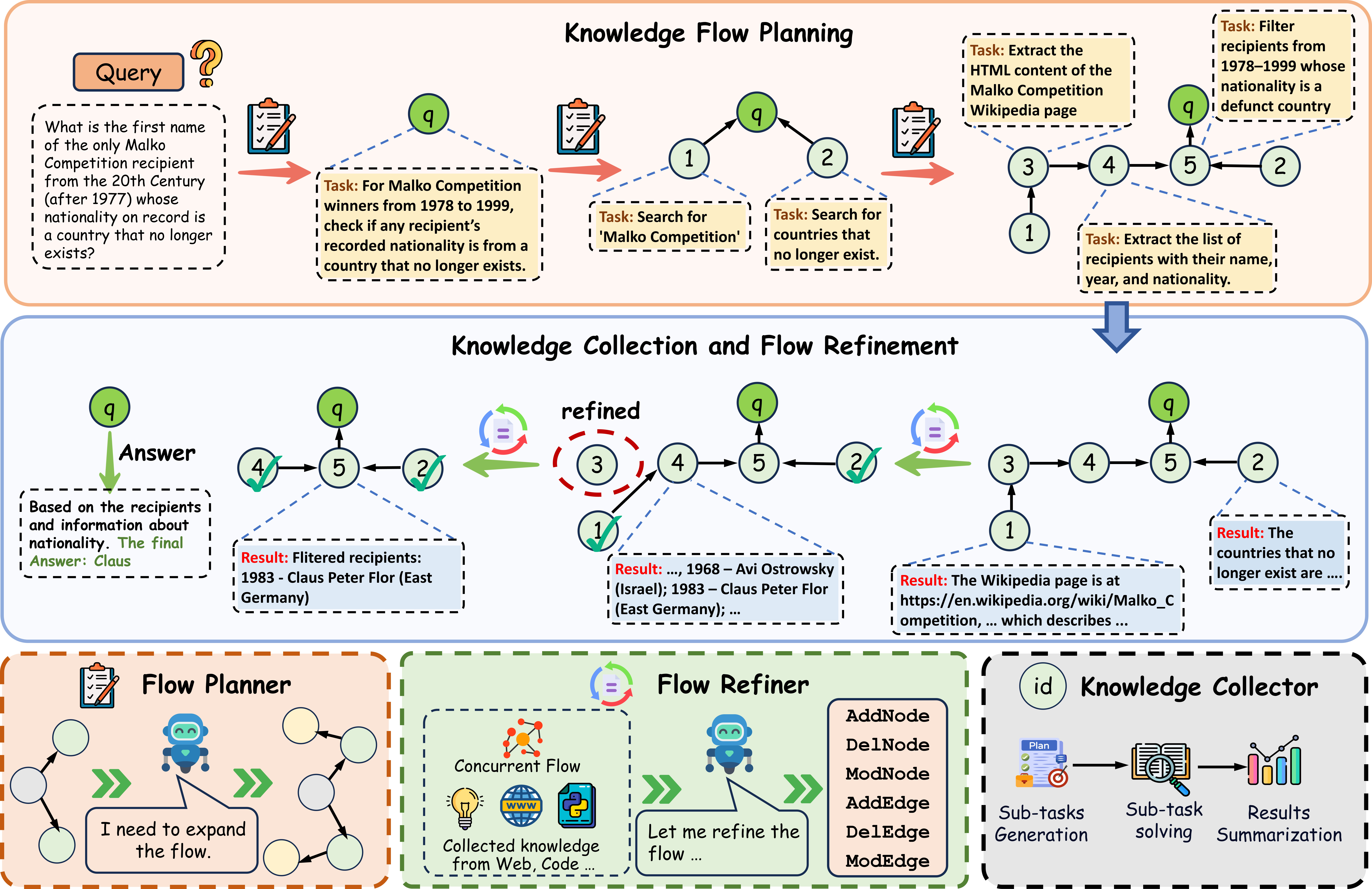} 
\end{center}
\vspace{-12pt}
\caption{Overview of {\ProjectName}. \textbf{Top part} illustrates the Knowledge Flow Planning process, where the Knowledge Flow Planner incrementally expands the structured knowledge flow.~\textbf{Middle part} depicts the iterative process of Knowledge Collection and Flow Refinement, where nodes are executed by the Knowledge Collector and the flow is dynamically adjusted by the Knowledge Flow Refiner based on newly acquired knowledge. \textbf{Lower part} highlights the three key components of {\ProjectName}—Flow Planner (left), Flow Refiner (center), and Knowledge Collector (right)—and their collaborative role in enabling systematic, adaptive, and efficient deep research.}
\vspace{-6pt}
\label{img:framework}
\end{figure}

{\ProjectName} is built upon a dynamic knowledge flow that enables structured and adaptive scientific research. As illustrated in Fig.~\ref{img:framework}, the system comprises three core components: \textbf{Knowledge Flow Planner}, which constructs high-quality knowledge flows tailored to the research objective; \textbf{Knowledge Collector}, which executes subtasks and enriches each node with relevant contextual information; and \textbf{Knowledge Flow Refiner}, which monitors progress and dynamically adjusts the flow based on intermediate outcomes and newly acquired knowledge. By enabling multiple agents to collaborate along this evolving flow, {\ProjectName} achieves systematic, scalable, and efficient problem solving. We begin by formalizing the concept of the structured knowledge flow, followed by detailed descriptions of Knowledge Flow Planner, Knowledge Collector, and Knowledge Flow Refiner.

\subsection{Structured Knowledge Flow}

Structured Knowledge Flow provides principled guidance for systematically organizing information to improve both the systematicity and effectiveness of deep research. Moreover, its graph-structured formulation enables flexible revision of local research objectives and knowledge dependencies when necessary.

A common practice in deep research agents is to address a user query \(q\) by assembling a strictly linear pipeline \(L(q)=[a_1,a_2,\dots,a_n]\) and executing it in order \(a_1\!\to\!a_2\!\to\!\cdots\!\to\!a_n\). The precedence relations are implicitly encoded by positional order, \textit{i.e.}, \(a_i \prec a_j \iff i<j\). Despite its procedural simplicity and ease of implementation, such a linear formalism fails to capture the inherently complex and non-linear dependencies of real-world research processes.

To better capture the complex structure of deep research reasoning processes, we adopt a directed acyclic graph
\(G=(V,E)\) to explicitly model both task dependencies and knowledge flow. Each node \(v_i\in V\) is a typed subtask node \(v_i=(t_i, d_i, s_i, c_i)\), where \(t_i\in \{search, solve, answer\}\) is the task type, $d_i$ is the task desciption, $s_i$ is the execution state of the node and $c_i$ is the resulting knwoledge context of the node if successfully executed. Each directed edge \(e_{ij}=(v_i,v_j,r_{ij})\in E\) specifies how the output of \(v_i\) conditions or constrains \(v_j\) using the relation \(r_{ij}\in R\), where $R$ is the set of relation types. This flow makes precedence among the nodes and supports parallel execution on independent branches, yielding a more expressive and verifiable substrate for Deep Research.

As an illustration, the following example describes a minimal graph in natural language form:

\begin{tcolorbox}[colback=gray!5, colframe=black, left=1mm, right=2mm, top=1mm, bottom=1mm]
{\footnotesize
\begin{verbatim}
{
 "nodes": [
  {"node_id": "n1", "task_type": "answer", "content": "<query>"},
  {"node_id": "n2", "task_type": "solve", "content": "<subtask>"},
  {"node_id": "n3", "task_type": "search", "content": "<subtask>"},
 ],
 "edges": [
  {"from": "n2", "to": "n1", "relationship": "solve subtask"},
  {"from": "n3", "to": "n1", "relationship": "provide information"},
 ]
}
\end{verbatim}
}
\end{tcolorbox}

Here, n1 represents an answering task defining the final research objectives, while n2 and n3 correspond to problem-solving and retrieval subtasks, respectively. The edges indicate dependencies: n2 supports the final answer, and n3’s retrieval is guided by the objective of n1.

By formalizing the research process in this manner, the agent is enabled not only to generate execution plans, but also to reason over the structural dependencies among subtasks, ensuring coherence and systematicity throughout the deep research workflow. Moreover, the graph-structured formulation allows multiple non-dependent nodes to be executed in parallel, and enables flexible revision of local research components without interfering with other ongoing processes, including the insertion of new nodes or the modification of suboptimal planning decisions when errors are identified, thereby enhancing the robustness of the overall system. These mechanisms will be further elaborated in the subsequent section on the construction of \ProjectName.

\subsection{Knowledge Flow Planner}

A high-quality Knowledge Flow is essential for the effective execution of complex research tasks. Rather than constructing the entire structure in a single step, which can lead to instability and reduced control, we employ a Knowledge Flow Planner process that incrementally initializes the flow.

Let $G_{t}^{init}=\{V_{t}^{init}, E_{t}^{init}\}$ be the flow in the $t$-th initialization iteration. Specifically, $G_{0}^{init}=\{\{v_{query}\}, \emptyset\}$ only contains the query node at the beginning. At each iteration $t$, an LLM planner examines the nodes in the current flow $G_t^{init}$ to identify those requiring further decomposition or additional context. For each node that requires decomposition, the planner generates a set of successor nodes representing sub-questions, intermediate reasoning steps, or supporting evidence for it. The corresponding dependency edges are added to the flow to maintain structural coherence and preserve logical consistency, which can be formulated as follows:
\begin{equation}
G_{t+1}^{init}=\{V_{t+1}^{init}, E_{t+1}^{init}\}=f_{\theta}^{expand}(G_t^{init}),
\end{equation}
where $f_{\theta}^{expand}(\cdot)$ is the trainable LLM planner, $V_{t+1}^{init} = V_t^{init}\cup{V_t^{add}}$ contains newly added nodes and $E_{t+1}^{init} = E_t^{init}\cup{E_t^{add}}$ contains newly introduced edges (dependencies) connecting nodes in $V_t^{add}$. This iterative expansion progressively extends the boundaries of the research and deepens the level of exploration within the knowledge flow. The process continues until $f_\theta^{expand}(\cdot)$ yields no additional nodes. Upon completion of the expansion phase, an initial flow $G_0=G_{T}^{init}$ is instantiated to support subsequent knowledge collector and flow refiner, where $T$ is the iteration steps in the flow expansion stage.

We curated a dataset of 10k examples to fine-tune a large model for the planner, which we term InternPlanner. Each data point is formatted as a dialogue: the input is a textual description of a flow, and the output is either (i) an updated flow obtained by expanding the current flow by one step, or (ii) the unchanged input flow, indicating the termination of the expansion. Further details about the dataset can be found in Appendix~\ref{appendix:data}.
 
After the initial planning of the knowledge flow, {\ProjectName} then enters another iterative loop of Knowledge Collector and Flow Refiner. This iteration continues until the original user query is successfully resolved. Knowledge Collector and Flow Refiner are described as follows.

\subsection{Knowledge Collector}

The Knowledge Collector aims at identifying the outermost executable nodes in the flow—those whose dependencies have all been resolved—and assigns each to an executor agent for processing. These agents, implemented as large language models equipped with tools, decompose the subtask into a sequential execution trajectory, iteratively reasoning and retrieving information to resolve the node. Available tools include web browsing, file downloading, and visual question answering, etc. A complete list of supported tools is provided in the Appendix~\ref{appendix:tools}.

After the execution of node $v_i$, its execution state $s_i$ (either success or failure) is updated. If the execution succeeds, the resulting knowledge—either retrieved or derived through reasoning—is distilled into a summarized knowledge context $c_i$, which serves as input for the subsequent execution of the nodes that depend on it. Formally,  given $G_t=(V_t, E_t)$ in the $t$-th Knowledge Collector and Flow Refiner iteration, the execution of node $v_i$ can be described as:
\begin{equation}
s_i, c_i = f^{exec}(t_i, d_i|\{c_j|(v_j \rightarrow v_i) \in E_t\}),
\end{equation}
where $t_i$ and $d_i$ are the task type and task description of node $v_i$, $f^{exec}(\cdot)$ is the LLM executor with tools depending on the context knowledge $\{c_j|(v_j \rightarrow v_i) \in E_t\}$. After the parallel execution of all the outermost executable nodes, a flow refinement will be conducted based on the newly obtained knowledge, which will be described in Sec.~\ref{sec:refiner}.

\subsection{Knowledge Flow Refiner}
\label{sec:refiner}

After completing the execution of nodes and updating the corresponding knowledge in the iteration, {\ProjectName} activates the Knowledge Flow Refiner to improve the structure of the flow. Leveraging the newly acquired knowledge, Knowledge Flow Refiner analyzes the current flow and identify potential structural adjustments, including the addition, removal, or modification of tasks and dependencies.
The goal of Knowledge Flow Refiner is to advance the research task in a reflective way and enhance execution efficiency.

The Knowledge Flow Refiner (achieved by an LLM) is prompted to utilize a set of predefined graph transformation operations to modify nodes and edges in the flow based on the knowledge context and execution states of the existing nodes. These operations include:

\begin{itemize}
    \item \textbf{Add Node (\texttt{AddNode})}: Introduce new nodes to capture missing sub-questions, intermediate reasoning steps, or evidence that were not anticipated in the initial flow.
    \item \textbf{Delete Node (\texttt{DelNode})}: Remove nodes that are redundant, irrelevant, or no longer necessary given the updated knowledge.
    \item \textbf{Modify Node (\texttt{ModNode})}: Modify the attributes of current nodes, especially the content of the sub-task. 
    \item \textbf{Add Edge (\texttt{AddEdge})}: Create new dependency edges to reflect newly discovered relationships between nodes.
    \item \textbf{Delete Edge (\texttt{DelEdge})}: Remove edges that represent incorrect, obsolete, or redundant dependencies, ensuring a more reasonable graph structure. 
    \item \textbf{Modify Edge (\texttt{ModEdge})}: Modify existing edges to correct dependency directions or improve the structure for more efficient execution.
\end{itemize}

Formally, $G_{t+1} = f^{refine}(\{V_t, E_t\})$, where $f^{refine}$ is an LLM that generates a sequence of graph transformation operations $\mathcal{O} = \{o_1, o_2, \dots, o_m\}$ and applies them on  $G_t=\{V_t, E_t\}$ to obtain the updated flow $G_{t+1}$. Through ongoing adjustments, {\ProjectName} achieves coherent and goal-directed reasoning.

\subsection{Conclusion Generation}
At the end of all iterations of knowledge collector and flow refiner, only the initial query node remains unexecuted. If the query is a straightforward scientific question that can be answered directly and simply, the query node will utilize the knowledge from its connected nodes to provide an immediate response. If the query requires generating a detailed scientific report, the final query node will aggregate knowledge from all nodes within the flow, perform a comprehensive reasoning process, and deliver a complete and thorough report. The details are explained in Appendix~\ref{app:summ}.

\section{Experiments}

To comprehensively assess the capabilities of \ProjectName, we conduct experiments on a diverse set of challenging benchmarks, ranging from general question answering to scientific deep research.

\subsection{Experiments Setup}

\textbf{Evaluation Benchmarks.} We conduct extensive experiments on four challenging benchmarks, including:
\begin{itemize}
    \item \textbf{GAIA}~\cite{mialon2023gaia}: a benchmark of real-world questions that require a set of fundamental abilities such as reasoning, multi-modality handling, web browsing, and generally tool-use proficiency. Our results are based on its 165-question validation set.
    \item \textbf{GPQA}~\citep{rein2024gpqa}: a benchmark of 448 multiple-choice questions across biology, chemistry, and physics, authored by domain experts to ensure depth and rigor, thereby providing a stringent evaluation of advanced reasoning and scientific knowledge. We use its 198-question GPQA-diamond subset for evaluation.
    \item \textbf{HLE}~\citep{phan2025humanity}: Humanity’s Last Exam is a multimodal benchmark consisting of 2,500 questions across mathematics, humanities, and natural sciences. Developed by subject experts, it provides a frontier-level test of academic competence where current LLMs still perform far below human experts.  
    \item \textbf{TRQA}~\citep{zhang2025origene}: a domain-specific benchmark for therapeutic target discovery. It covers fundamental biology, disease biology, pharmacology, and clinical medicine, providing a systematic evaluation framework for biomedical research agents.  We use its 172-question TRQA-lit subset for evaluation.
\end{itemize}

\textbf{Methods of Comparison.} To validate the effectiveness of \ProjectName, we compare \ProjectName~ on GAIA, GPQA, HLE and TRQA against cutting-edge large language models including Qwen3 series model~\citep{yang2025qwen3}, Intern-S1~\citep{bai2025intern}, Deepseek-R1~\citep{guo2025deepseek}, GPT-o4-mini, and GPT-5~\citep{chatgpt2025}, and some state-of-the-art deep research agent, including proprietary approaches OpenAI-DR (Deep Research)~\citep{openai2025deepresearch}, Gemini-DR~\citep{google2024geminidr} and Manus~\citep{manus2025}, 
leading react agentic models MiroThinker~\citep{2025mirothinker}, Tongyi-DR~\citep{team2025tongyi}, WebShaper~\citep{tao2025webshaper}, WebDancer~\citep{wu2025webdancer},
and open-source frameworks MiroFlow~\citep{2025mirothinker}, OWL~\citep{hu2025owl}, X-Masters~\citep{chai2025scimaster}, AWorld~\citep{yu2025aworld}, and OriGene~\citep{zhang2025origene}. 
In the experiments, we utilize GPT-o4-mini to serve as the Knowledge Flow Planner, Knowledge Collector and Knowledge Flow Refiner in our workflow.

\begin{table}[t]
\vspace{-4pt}
\centering
\caption{Performance comparison on GAIA, GPQA-diamond and HLE benchmarks. The best results are \textbf{bolded} and the second best results are \underline{underlined}. Results not reported in the original papers are denoted as `` - ". \textbf{Note that} some approaches, such as MiroThinker and TongyiDR, can only report results on the HLE text-only subset as these approaches lack multimodal ability.}
\label{tab:main_results}
\vspace{-6pt}

\setlength{\tabcolsep}{3.2pt}
\resizebox{0.99\linewidth}{!}{
\begin{tabular}{lccccccccccc}
% \toprule
\hline
                &  & \multicolumn{4}{|c|}{\textbf{GAIA val}}                          & \multicolumn{4}{c|}{\textbf{GPQA-diamond}}            & \multicolumn{2}{c}{\textbf{HLE}}                                   \\ 
% \midrule
\hline
\multicolumn{1}{c}{\textbf{Method}}  &\multicolumn{1}{c|}{\textbf{Base Model}}                 & Level 1 & Level 2 & Level 3 & \multicolumn{1}{c|}{Avg.} & Bio & Chem & Phys & \multicolumn{1}{c|}{Avg.} & \begin{tabular}[c]{@{}c@{}}text only\end{tabular} & All \\ 
\hline
% \midrule
\rowcolor{pink!20}
\multicolumn{12}{c}{\textit{\textbf{No Agency}}}                                                                                                                                                                         \\ 
% \midrule
\hline
\multicolumn{1}{c}{Qwen-3-8B}  &\multicolumn{1}{c|}{-}         &    11.32     &   2.32      &   0.00      &   \multicolumn{1}{c|}{4.85}    &  -   &  -    &  -    & \multicolumn{1}{c|}{44.44}    & -                                                    &  -   \\
\multicolumn{1}{c}{Qwen3-32B}  &\multicolumn{1}{c|}{-}          &     13.21    &   3.49      &    3.84     & \multicolumn{1}{c|}{6.67}    &   -  &  -    &  -    & \multicolumn{1}{c|}{49.49}    &      -                                               &  -   \\
\multicolumn{1}{c}{Qwen3-235B}  &\multicolumn{1}{c|}{-}          &    15.09     &   3.49      &     3.84    & \multicolumn{1}{c|}{7.27}    & -    &  -    &  -    & \multicolumn{1}{c|}{47.47}    &   9.18                                                  & 8.60    \\
\multicolumn{1}{c}{Intern-S1} &\multicolumn{1}{c|}{-}           &28.30    &   9.30      &    7.69     & \multicolumn{1}{c|}{15.15}    &  \textbf{89.47}   &  59.49    &  93.02    & \multicolumn{1}{c|}{78.26}    & 8.90                                                    & 8.30    \\
\multicolumn{1}{c}{Deepseek-R1} &\multicolumn{1}{c|}{-}         &   33.96      &   13.95      &   3.84      & \multicolumn{1}{c|}{18.78}    &  63.16   &  \underline{76.34}    &  91.86    & \multicolumn{1}{c|}{82.32}    & 8.60                                                    &-     \\
\multicolumn{1}{c}{o4-mini}  &\multicolumn{1}{c|}{-}            &   28.30      &   12.79      &    7.69     & \multicolumn{1}{c|}{16.97}    &  78.95   &  63.44    &  94.19    & \multicolumn{1}{c|}{78.28}    &14.50                                                     & 14.28    \\
\multicolumn{1}{c}{GPT-5}   &\multicolumn{1}{c|}{-}             &  -       &   -      &     -    & \multicolumn{1}{c|}{-}    & \underline{84.21}    & \underline{76.34}     & \underline{95.35}     & \multicolumn{1}{c|}{\underline{85.35}}    & 25.85                                                    & 24.76    \\ 
% \midrule
\hline
\rowcolor{orange!10}  
\multicolumn{12}{c}{\textit{\textbf{Close-sourced Agentic Framework}}}    \\ 
\hline
% \midrule
\multicolumn{1}{c}{OpenAI DR}  &\multicolumn{1}{c|}{-}          & 74.29        &  69.06       &  47.60       & \multicolumn{1}{c|}{67.36}    &   -  &   -   &  -    & \multicolumn{1}{c|}{-}    & -                                                    & 26.60    \\
\multicolumn{1}{c}{Manus}  &\multicolumn{1}{c|}{-}               & 86.50        & 70.10        & \underline{57.70}        & \multicolumn{1}{c|}{73.30}    &-     & -     &  -    & \multicolumn{1}{c|}{-}    &-                                                     & -    \\
\multicolumn{1}{c}{Gemini DR} &\multicolumn{1}{c|}{-}  & -        & -        &   -      & \multicolumn{1}{c|}{-}    & -    &  -    &   -   & \multicolumn{1}{c|}{-}    & -                                                    & 26.90    \\ 
% \midrule
\hline
\rowcolor{green!10}
\multicolumn{12}{c}{\textit{\textbf{React Agentic Model}}}   \\ 
\hline
% \midrule
\multicolumn{1}{c}{\pdfdiff{WebDancer}}   &\multicolumn{1}{c|}{\pdfdiff{QwQ-32B}}           & \pdfdiff{61.5}        & \pdfdiff{50.0}        &  \pdfdiff{25.0}       & \multicolumn{1}{c|}{\pdfdiff{51.5}}    &  -   & -     &  -    & \multicolumn{1}{c|}{-}    &  -                                                   & -    \\
\multicolumn{1}{c}{\pdfdiff{WebShaper}}   &\multicolumn{1}{c|}{\pdfdiff{Qwen2.5-72B}}           & \pdfdiff{69.2 }       & \pdfdiff{63.4 }       & \pdfdiff{ 16.6}      & \multicolumn{1}{c|}{\pdfdiff{60.1}}      & \pdfdiff{47.37}     & \pdfdiff{ 52.69 }   &  \pdfdiff{81.40} & \multicolumn{1}{c|}{\pdfdiff{64.65}}    &  -                                                  & -    \\
\multicolumn{1}{c}{\pdfdiff{MiroThinker}}   &\multicolumn{1}{c|}{\pdfdiff{Qwen2.5-72B}}           & -       & -       & -     &\multicolumn{1}{c|}{\underline{81.9}}     &\underline{84.21}     & 75.27   &  \underline{95.35} &\multicolumn{1}{c|}{84.85}  &  \textbf{37.7}                                                  & -    \\
\multicolumn{1}{c}{\pdfdiff{Tongyi DR}}   &\multicolumn{1}{c|}{\pdfdiff{Qwen3-30B-A3B}}           & -      & -       &  -     & \multicolumn{1}{c|}{70.9}      & 78.95     & 67.74  &  \underline{95.35} & \multicolumn{1}{c|}{80.30}  &  32.9                                                  & -    \\
% \midrule
\hline
\rowcolor{blue!10}
\multicolumn{12}{c}{\textit{\textbf{Open-sourced Agentic Framework}}}                                                                                                                                                            \\ 
\hline
% \midrule
\multicolumn{1}{c}{MiroFlow}   &\multicolumn{1}{c|}{Claude-3.7}           & -        & -        &  -       & \multicolumn{1}{c|}{74.50}   &  -   & -     &  -    & \multicolumn{1}{c|}{-}    &  29.50                                                   & 27.20    \\
\multicolumn{1}{c}{OWL}   &\multicolumn{1}{c|}{Gemini-2.5-Pro}               & 84.90        & 68.60        & 42.30        & \multicolumn{1}{c|}{69.70}    &\pdfdiff{57.89}     & \pdfdiff{61.29}     & \pdfdiff{86.05}     & \multicolumn{1}{c|}{\pdfdiff{71.72}}    &  -                                                   & -    \\
\multicolumn{1}{c}{X-Masters} &\multicolumn{1}{c|}{Deepseek-R1}           & -        &-         & -        & \multicolumn{1}{c|}{-}    & 78.95    &  68.82    &  94.19    & \multicolumn{1}{c|}{80.81}    & 32.10                                                    & \underline{27.72}    \\
\multicolumn{1}{c}{JoyAgent}   &\multicolumn{1}{c|}{\pdfdiff{Claude-4}}               & \pdfdiff{86.79}        & \underline{77.91}        & \pdfdiff{42.31}        & \multicolumn{1}{c|}{75.15}    &78.95     & 65.59     &  91.86    & \multicolumn{1}{c|}{77.27}    &  -                                                   & -    \\ 
\multicolumn{1}{c}{\pdfdiff{AWorld}}   &\multicolumn{1}{c|}{\pdfdiff{Gemini-2.5-Pro}}               & \underline{88.68}        & \underline{77.91}        & 53.85        & \multicolumn{1}{c|}{77.58}    &73.68     & 66.67     & 93.02     & \multicolumn{1}{c|}{78.79}    &  -                                                   & -    \\
% \midrule
\hline
\rowcolor{mycolor!10}  
\multicolumn{12}{c}{\textit{\textbf{\ProjectName}}}  \\ 
\hline
% \midrule
\multicolumn{1}{c}{\ProjectName}&\multicolumn{1}{c|}{Qwen3-235B}       &    69.81     &   60.47      &     30.77    & \multicolumn{1}{c|}{58.79}    & 63.16   &58.06      & 75.58     & \multicolumn{1}{c|}{66.16}    & 15.04                                                    &14.84     \\ 
\multicolumn{1}{c}{\ProjectName}&\multicolumn{1}{c|}{o4-mini}       &  \textbf{92.45}       &   \textbf{82.56}      &  \textbf{61.54}       & \multicolumn{1}{c|}{\textbf{82.42}}    & \underline{84.21}    & \textbf{79.57}     & \textbf{96.51}     & \multicolumn{1}{c|}{\textbf{87.37}}    & \underline{36.10}                                                    & \textbf{34.52}    \\ 
% \bottomrule
\hline
\end{tabular}
}
\end{table}

\subsection{Experiment Results}
\begin{figure}[h]
\vspace{-6pt}
\begin{center}
\includegraphics[width=0.5\textwidth]{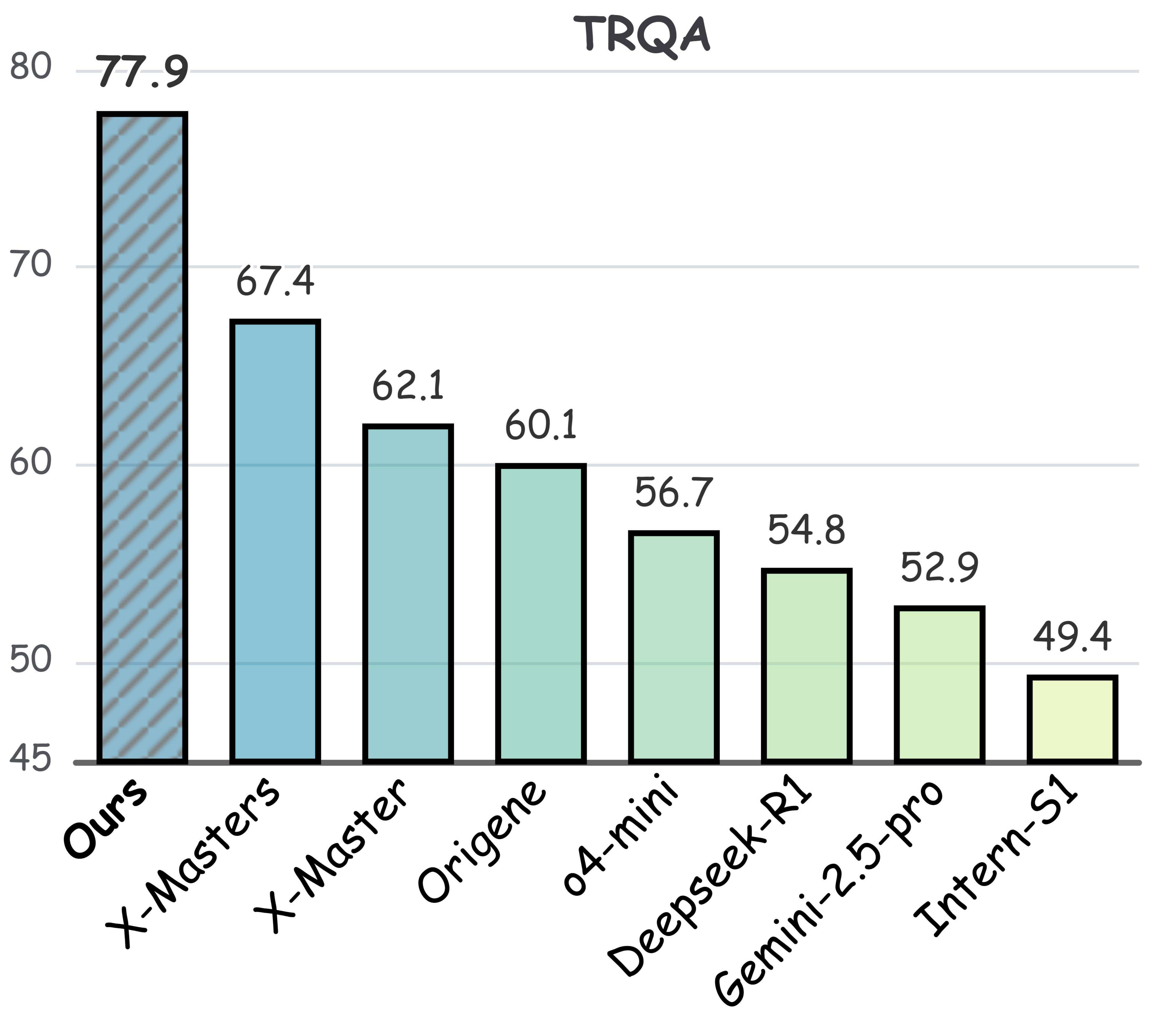} 
\end{center}
\vspace{-12pt}
\caption{Performance on the TRQA benchmark. {\ProjectName} (Ours) significantly outperforms previous works.}
\vspace{-2pt}
\label{fig:trqa}
\end{figure}

Table~\ref{tab:main_results} and Fig.~\ref{fig:trqa} present the performance of \ProjectName~and its counterparts on GAIA, GPQA, HLE, and TRQA. \ProjectName ~consistently achieves competitive results across all benchmarks without training an additional base model, validating the effectiveness of its systematic design.

\subsubsection{Agentic Benchmark}

\textbf{\ProjectName~ achieves state-of-the-art performance among agentic systems.} On GAIA (Table~\ref{tab:main_results}), \ProjectName~ (o4-mini) outperforms both closed-source Manus (73.30\%) and leading open-source agentic models Mirothinker (81.9\%) and Tongyi-DR (70.9\%), even though they are specifically trained and evaluated only on the GAIA text-only subset. \ProjectName~ also shows strong robustness on Level 3 questions (61.54\%). These results indicate that its iterative workflow combining knowledge planning, collection, and refinement is particularly effective for multi-hop and compositional reasoning. Its clear advantage over systems like OpenAI DR and MiroFlow further underscores the impact of structured and dynamic workflow design.

\begin{table}[t]
\vspace{-4pt}
\caption{Ablation study on the impact of structured planning and refinement. We compare the workflow with conventional sequential planner, the flow planner, and the flow refiner. A checkmark (\checkmark) indicates the component is used. Results are reported on GAIA and GPQA.}
\label{tab:ab_planner_type}
\centering
\vspace{-6pt}
\resizebox{\linewidth}{!}{
\begin{tabular}{ccc|cccc|cccc}
% \toprule
\hline
& &  & \multicolumn{4}{c|}{\textbf{GAIA}} & \multicolumn{4}{c}{\textbf{GPQA}} \\ 
\hline
% \midrule
\textbf{Sequential Planner} & \textbf{Flow Planner} & \textbf{Refiner} & Level 1 & Level 2 & Level 3 & Avg & Bio & Chem & Phys & Avg \\ 
\hline
% \midrule
 \checkmark& -- & -- & 67.92 & 55.81 & 23.07 & 55.76 & 57.89 & 54.84 & 88.37 & 71.21 \\ 
 --& \checkmark & -- & 73.58 & 63.95 & 30.77 & 61.82 & 57.89 & 59.14 & 89.53 & 73.74 \\ 
-- & \checkmark & \checkmark & \textbf{92.45} & \textbf{82.56} & \textbf{61.54} & \textbf{82.42} & \textbf{84.21} & \textbf{79.57} & \textbf{96.51} & \textbf{87.37} \\ 
\hline
% \bottomrule
\end{tabular}
}
\end{table}

\textbf{Agentic systems consistently outperform pure LLMs on complex reasoning tasks.} Larger base models like the Qwen series benefit from greater internal knowledge, but remain limited without structured reasoning. Even models fine-tuned for scientific reasoning, such as Intern-S1, lag behind agentic approaches. For instance, \ProjectName~ with o4-mini achieves a score of 82.42\% on GAIA, far surpassing the same model (o4-mini) without agency (16.97\%), highlighting that structured task decomposition and flow-based execution are more critical than model size alone.

\subsubsection{Multi-Disciplinary Research and Question Answering}

Although scientific benchmarks are generally knowledge-intensive and favor LLMs, \ProjectName~ still achieves competitive performance.

\textbf{\ProjectName~ effectively acquires domain-specific knowledge through Knowledge Flow.} On the GPQA-diamond benchmark, \ProjectName~ (o4-mini) achieves 87.37\% average accuracy in Biology, Chemistry, and Physics—outperforming GPT-5 and Intern-S1. This underscores the advantage of dynamic retrieval in accessing context-relevant knowledge, which enables more accurate and flexible scientific reasoning than relying solely on static information.

\textbf{General-purpose tools guided by Knowledge Flow can outperform specialized systems.} On the HLE benchmark, \ProjectName~ (o4-mini) achieves the highest accuracy among training-free methods at 34.52\%, demonstrating performance comparable to leading trained agentic models such as MiroThinker (37.7\%) and Tongyi-DR (32.9\%) on the text-only subset, as well as closed-source systems including OpenAI DR (26.60\%) and Gemini Deep Research (26.90\%). On TRQA, \ProjectName~ reaches 77.9\%, outperforming domain-specific agent Origene (60.1\%) and scientific multi-modal model Intern-S1 (49.4\%). These results show that a well-structured general-purpose agent system can effectively tackle complicated scientific and cross-domain tasks.

\subsection{Ablation Studies}

\textbf{Ablation on key components.} We conduct ablation studies on two critical components of {\ProjectName}: the Knowledge Flow Planner and the Knowledge Flow Refiner. As shown in Table~\ref{tab:ab_planner_type}, replacing conventional sequential planner reasoning with our structured Knowledge Flow leads to substantial performance improvements, with gains of 6.06\% on GAIA and 2.53\% on GPQA. This highlights its effectiveness in capturing complex task dependencies and enhancing problem-solving capabilities. Moreover, incorporating the Flow Refiner yields further notable improvements, indicating that dynamic flow refinement enables more flexible task adaptation and strengthens the agent’s overall research competence.

\begin{table}[t]
\vspace{-4pt}
\caption{Ablation study on the planner model. We compare various flow planners, including the Qwen3 series and our finetuned InternPlanner. Results are reported on the GAIA benchmark.
}
\label{tab:ab_planner}
\centering
\begin{tabular}{c|cccc}
\hline
                  & \multicolumn{4}{c}{\textbf{GAIA}}    \\ \hline
\textbf{Planner}           & Level 1 & Level 2 & Level 3 & Avg  \\ \hline
Qwen-3-8B         &  58.49       &  46.51       & 11.54        &  44.85      \\
InternPlanner-8B (ours)  & 70.25        &  67.44       &  34.61       &  66.06      \\
\arrayrulecolor{lightgray}
\hline
\arrayrulecolor{black}
Qwen-3-32B        &  77.36       &   67.44      &  30.77       &  64.81      \\
InternPlanner-32B (ours) &  \textbf{84.91}       &   \textbf{70.93}      &  \textbf{42.31}       &  \textbf{70.91}      \\ 
\hline
\end{tabular}
\end{table}

\textbf{Ablation on trained flow planner.}
We conduct experiments with different Flow planners to assess their impact on {\ProjectName}’s performance. As shown in Table~\ref{tab:ab_planner}, comparing Qwen3-8B and Qwen3-32B reveals a clear trend: stronger base models produce higher-quality Knowledge Flows, which in turn lead to better overall performance. Moreover, our InternPlanner, finetuned from Qwen3-8B and Qwen3-32B, consistently outperforms their original counterparts, demonstrating both the critical role of the planner and the effectiveness of our training strategy.

\subsection{Case Study and Visualization}
\begin{figure}[h]
\vspace{-6pt}
\begin{center}
\includegraphics[width=\textwidth]{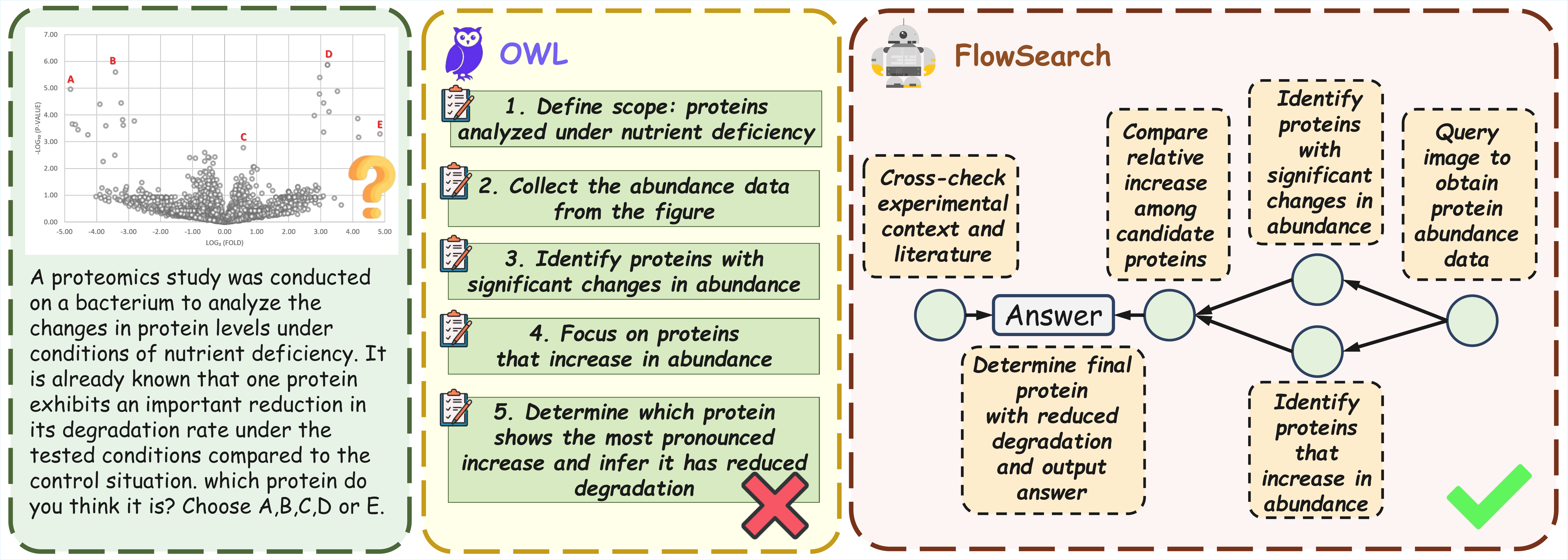} 
\end{center}
\vspace{-12pt}
\caption{Case study comparing the conventional deep research framework OWL with our {\ProjectName} on a scientific question.}
\vspace{-2pt}
\label{fig:case_study}
\end{figure}

Fig.~\ref{fig:case_study} illustrates the contrast between our knowledge-flow-based {\ProjectName} and the conventional sequential planning paradigm, represented by OWL~\citep{hu2025owl}, in addressing a scientific question. As shown in the figure, OWL decomposes the query into a linear sequence of subtasks—such as understanding, information collection, and identification—that are executed in order. While this pipeline is straightforward, it lacks mechanisms to preserve and integrate intermediate insights, which leads to the dilution of valuable evidence as the chain grows longer.

In comparison, {\ProjectName} constructs a structured knowledge flow directly from the user query, explicitly modeling dependencies between subtasks—for instance, asking a question about an image only after extracting information from it. Each node both executes its designated operation and summarizes its outcome, passing structured intermediate results to subsequent steps along the flow. This design enables selective reuse of prior knowledge, limits the propagation of irrelevant information, and preserves critical evidence throughout the reasoning process.

\section{Conclusion}
In this work, we have presented \textbf{{\ProjectName}}, a multi-agent deep research system built on a dynamic structured knowledge flow. By explicitly modeling dependencies among subproblems and key concepts, the system enables both deep reasoning within local regions of the knowledge flow and coherent knowledge propagation at a global level. The dynamic flow framework allows {\ProjectName} to iteratively plan, expand, and refine research workflows, supporting hierarchical task decomposition, parallel exploration, and adaptive strategy adjustment based on intermediate findings. These results highlight the effectiveness of combining structured knowledge flow planning with multi-agent orchestration, suggesting that such frameworks offer a promising direction for building autonomous, reflective, and scalable systems capable of tackling complex scientific research tasks.
%\clearpage

\begingroup
\sloppy
\printbibliography[heading=bibintoc]
% \bibliography{references}
% \input{main.bbl}
\endgroup

\clearpage
\appendix

\section{Tools of Knowledge Collector}
\label{appendix:tools}
We provide a set of tool wrappers used by the Knowledge Collector.  Each tool is designed with concurrent safety, creating independent toolkit instances to avoid state conflicts.  Table~\ref{tab:exec-tools} summarizes the available tools.

\begin{table}[h]
\centering
\caption{The tools in Knowledge Collector}
\label{tab:exec-tools}
\resizebox{1.0\linewidth}{!}{
\begin{tabular}{p{5cm} p{11cm}}
\toprule
\textbf{Tool} & \textbf{Purpose} \\
\midrule
\texttt{search\_google} & Use Google search engine to search information for the given query \\
\arrayrulecolor{lightgray}
\midrule
\texttt{search\_wiki} & Search the entity in WikiPedia and return the summary of the
            required page, containing factual information about
            the given entity \\
\midrule
\texttt{search\_wiki\_revision} & Search Wikipedia to get the latest Wikipedia revision *at or before* the end of the given (year, month)\\ 
\midrule
\texttt{search\_archived\_webpage} & Given a url, search the wayback machine and returns the archived version of the url for a given date \\
\midrule
\texttt{extract\_document\_content} & Extract the content of a given local document and return the processed text. It can process various types of documents, including text, image, table, audio, video, zip, json, xml, pdf, py etc \\
\midrule
\texttt{extract\_url\_content} &  Extract the html content of a given url and return the processed text \\
\midrule
\texttt{ask\_question\_about\_image} & Answer image questions with optional custom instructions \\
\midrule
\texttt{ask\_question\_about\_audio} & Ask a question about the audio and get the answer using
        multimodal model \\
\midrule
\texttt{ask\_question\_about\_video} & Ask a question about the video using Gemini multimodal capabilities \\
\midrule
\texttt{download\_media\_from\_url} &     Download any given URL (image, video, audio, document, or webpage)\\
\midrule
\texttt{execute\_code} & Execute a given code snippet \\
\midrule
\texttt{browse\_url} & A powerful toolkit which can simulate the browser interaction to solve the task which needs multi-step actions \\
\midrule
\texttt{ocr2text} & OCR the image and return the text \\
\arrayrulecolor{black}
\bottomrule
\end{tabular}
}
\end{table}

\section{Summarizer}

A summarizer has been developed to generate conclusions for the answer node, featuring two modes of operation.

\textbf{Question-answering tasks}:
When the objective is to answer a specific question, the task usually requires a strict logical progression of reasoning. In such cases, the later nodes in the execution graph—particularly solve nodes—tend to encapsulate the reasoning steps that are most directly related to the final answer. By contrast, earlier nodes such as search nodes often contain intermediate evidence or auxiliary information that, while necessary for the reasoning process, does not itself contribute to the correctness or clarity of the final response. To enhance efficiency and reduce noise, the summarizer in this mode selectively incorporates only the dependent predecessor nodes of the final answer. This targeted approach ensures that the summary remains focused, concise, and aligned with the logical chain that directly supports the solution. The main benefit is an improvement in answer precision and interpretability, as irrelevant or redundant details are filtered out.

\textbf{Report-generation tasks}:
By contrast, when the task involves producing a comprehensive report, the goal is not merely accuracy but also coverage and richness. In this context, limiting the summarizer to dependent nodes would risk omitting potentially valuable background, context, or supporting evidence. Therefore, for report generation, the entire execution graph produced by {\ProjectName} is passed to the summarizer. This design allows the system to synthesize information from all nodes—including search, solve, and answer stages—so that the final report captures not only the core reasoning steps but also the broader landscape of evidence. The benefit of this approach is that the generated report provides a more holistic view of the research process, offering readers both the conclusions and the supporting context, which increases transparency, interpretability, and informational richness.

\textbf{Advantages of the dual-mode design}:
This bifurcated summarization strategy balances efficiency with completeness. For question answering, it minimizes cognitive load and reduces error propagation by concentrating only on essential reasoning chains. For report writing, it maximizes informativeness and ensures that potentially useful evidence is not prematurely discarded. Together, these modes enable {\ProjectName} to flexibly support both precise problem-solving and broad knowledge synthesis, depending on the user’s research goals.

% \section{Training of the flow planner}

\section{Dataset and Training}
\label{appendix:data}
We employ knowledge distillation from GPT-O4-mini to train our InternPlanner. Specifically, for the training data, we collect a set of Wikipedia entries that inherently exhibit entity dependencies. These dependencies are extracted and organized into structured entity graphs. The entity graphs are then fed into O4-mini, which generates questions for each node based on its corresponding dependencies and subsequently integrates these questions into a single summary question, serving as the user query in our dataset. For the graph generation process, the obtained entity graphs are directly converted into a natural language representation according to our predefined format, which is included as the assistant output in the labeled dataset. A detailed description of the data format is provided in Box~\ref{box:dataformat}.

During training, the labeled data is decomposed into multiple single-turn question-answer pairs, which are then used to fine-tune Qwen3-series models via supervised fine-tuning (SFT).

\begin{tcolorbox}[title=D. Data Format, breakable, colback=gray!5, colframe=black, label={box:dataformat}]
\begin{lstlisting}[breaklines=true,basicstyle=\ttfamily\small]
{
  "messages": [
    {
      "role": "user",
      "content": "You are a graph planner agent. Your task is to decompose any user question into a logical graph of tasks, and iteratively refine the graph when node knowledge becomes available.  

        ### Example Input Graph
        {
          "nodes": [
            {"node_id": "n1", "type": "answer", "task": "Explain why sugar-free drinks can still contain carbohydrates"}
          ],
          "edges": []
        }
        
        Input graph:
        <generated_input_graph>"
    },
    {
      "role": "assistant",
      "content": "<generated_output_graph>"
    }
  ]
}
\end{lstlisting}
\end{tcolorbox}

\section{Additional Case Studies}
\subsection{Question Answering}

Below we choose one question from GAIA level 2 to show our execution logic. The content of the question is \textbf{``What is the latest chronological year date in the image from the first citation of Carl Nebel’s Wikipedia page (Aug 2023)?''}. This question is inherently \emph{logic-intensive}: the answer is not present on the query page but must be derived through a chained, evidence-preserving procedure.

Our {\ProjectName} solves this question step by step, in a very logical order. Starting from the revision-resolved entry point, the planner instantiated a dependency-aware, tool-grounded pipeline: \texttt{n1} resolves the August~2023 Wikipedia revision; \texttt{n2} extracts the first citation URL; \texttt{n3} fetches the citation page HTML; \texttt{n8} identifies and downloads the first in-article image; \texttt{n6} performs OCR over the downloaded image; and \texttt{n7} parses four-digit years to determine the latest date. This design operationalizes a search–extract–process–analyze flow where each step produces verifiable intermediate artifacts (URLs, HTML snapshots, files, OCR text) that can be audited, cached, and reused. By encoding dependencies in a graph, {\ProjectName} enables deterministic, provenance-preserving re-execution, isolates errors to specific nodes, and supports targeted recovery without rerunning the entire pipeline—yielding stronger reproducibility, interpretability, and multi-tool coordination than monolithic, end-to-end prompting. Table~\ref{tab:case-study} shows every node in our {\ProjectName} graph system, including task content, type, status and its output(part).

\noindent Since this question requires each step to rely strictly on the results of the previous one, the overall process forms a linear execution flow. The execution followed a strict topological ordering:
\begin{quote}
\centering
n1 $\rightarrow$ n2 $\rightarrow$ n3 $\rightarrow$ n8 $\rightarrow$ n6 $\rightarrow$ n7 $\rightarrow$ task
\end{quote}

\begin{table}[ht]
\centering
\caption{Execution Trace of {\ProjectName}(Question Answering Case Study)}
\label{tab:case-study}
\small
% \resizebox{\linewidth}{!}{
\begin{tabular}{p{0.6cm} p{3.8cm} p{1.2cm} p{1.2cm} p{4.8cm}}
\toprule
\textbf{Node} & \textbf{Task} & \textbf{Type} & \textbf{Status} & \textbf{Output(Part)} \\
\midrule
n1 & Use \texttt{search\_wiki\_revision} to get Carl Nebel Wikipedia revision (Aug 2023) & search & Success & The Carl Nebel Wikipedia page revision dated 2023-08-05T13:53:28Z carries oldid 1168855983 and is accessible at \url{https://en.wikipedia.org/w/index.php?title=Carl_Nebel&oldid=1168855983} \\
\midrule
n2 & Open Carl Nebel revision and extract first citation URL from References & solve & Success & The first citation in the References section is 'Thieme–Becker', entry “Nebel, Carl,” with URL \url{https://de.wikipedia.org/wiki/Thieme-Becker} \\
\midrule
n3 & Use \texttt{extract\_url\_content} to fetch citation page HTML & search & Success & Extracted the HTML content of the German Wikipedia page at \url{https://de.wikipedia.org/wiki/Thieme-Becker}, which presents Thieme-Becker as a German biographical dictionary of artists... \\
\midrule
n8 & Extract first image from citation page and download locally & search & Success & The first image in the ‘Thieme‐Becker’ article body has source URL \url{https://upload.wikimedia.org/wikipedia/commons/thumb/c/c5/Perwanger%2C_Chr...} \\
\midrule
n6 & Apply \texttt{ocr2text} on downloaded image & solve & Success & The OCR2Text tool returned success and extracted the following lines of text: Pervinquiere-Pescatori; Pervinquiere, Henri Baron, Tiermaler,; Perz, Michael, Stukkator, tatig 1701 im; 1883 Poitiers; Bild im Mus. ebda.; Rathaus zu Landsberg a. Lech... \\
\midrule
n7 & Parse OCR text to extract all 4-digit years and find latest & solve & Success & From the OCR‐extracted text, the unique four‐digit years identified are [1558, 1577, 1610, 1645, ... 1913, 1915, 1924, 1925, 1927], and the latest chronological year among these is 1927.
 \\
\midrule
task & What is the latest chronological year date in the image from the first citation of Carl Nebel’s Wikipedia page (Aug 2023)? & answer & Success & 1927 \\
\bottomrule
\end{tabular}
% }

\end{table}

\subsection{Report Generation}
\label{app:summ}
The following report was generated by our {\ProjectName} system to answer the query \textbf{``Help me research the latest progress in multi-agent AI scientists in 2025''}. The planner decomposed the original research query into a set of interconnected subtasks, enabling systematic information gathering, reasoning, and synthesis. The resulting execution graph consisted of 7 nodes spanning three major categories—search, solve, and answer—which together captured the full problem-solving workflow. Table~\ref{tab:case-nodes} provides a structured overview of all nodes and their corresponding roles within the pipeline. Importantly, in the case of report generation, the graph structure highlights its advantages even more clearly, since the dependencies across different report sections are relatively weak, allowing parallel execution to be fully leveraged. Therefore, we can generate the report below in 10 minutes. An example output report is shown below.

\noindent The execution followed a topological ordering, where nodes grouped together indicate parallel execution:

\begin{quote}
\centering
n3, n4s, n7 $\rightarrow$ n2, n4, n6 $\rightarrow$ task
\end{quote}

\begin{table}[h]
\centering
\caption{Execution Trace of {\ProjectName}(Report Generation Case Study)}
\label{tab:case-nodes}
\begin{tabular}{p{1.6cm} p{1.6cm} p{7cm}}
\toprule
\textbf{Node ID} & \textbf{Type} & \textbf{Task Description} \\
\midrule
task  & answer & Help me research the latest progress in multi-agent AI scientists in 2025.  \\
\midrule
n2    & solve  & [Background \& Methods] Synthesize definitions, scope, historical context, and classify core methods \\
\midrule
n3    & search & [Background \& Methods] Collect definitions, seminal works, and representative methods \\
\midrule
n4    & solve  & [Datasets/Applications] Summarize datasets, benchmarks, evaluation metrics, and applications \\
\midrule
n4s   & search & [Datasets/Applications] Collect datasets, benchmarks, evaluation results, and application examples \\
\midrule
n6    & solve  & [Challenges \& Future Work] Analyze challenges, limitations, and outline future directions \\
\midrule
n7    & search & [Challenges \& Future Work] Collect discussions of current challenges, limitations, and future outlook \\
\bottomrule
\end{tabular}
\end{table}

\clearpage

\textbf{The following content is an example of a report generated by {\ProjectName}:}

% ---------- Local styling just for this appendix ----------
\begingroup

\setlength{\parskip}{4pt}
\setlength{\parindent}{0pt}

\small                                % slightly smaller text for long prose
% Compact itemize/enumerate (optional; comment out if you don't use lists)
\makeatletter
\def\@listi{\leftmargin\leftmargini \parsep 2pt \topsep 4pt \itemsep 2pt}
\let\@listI\@listi
\makeatother
\vspace{2mm}\hrule\vspace{3mm}

\begin{figure*}[b]
\vspace{-100pt}
\begin{center}
\fbox{
\includegraphics[page=1,width=0.9\textwidth]{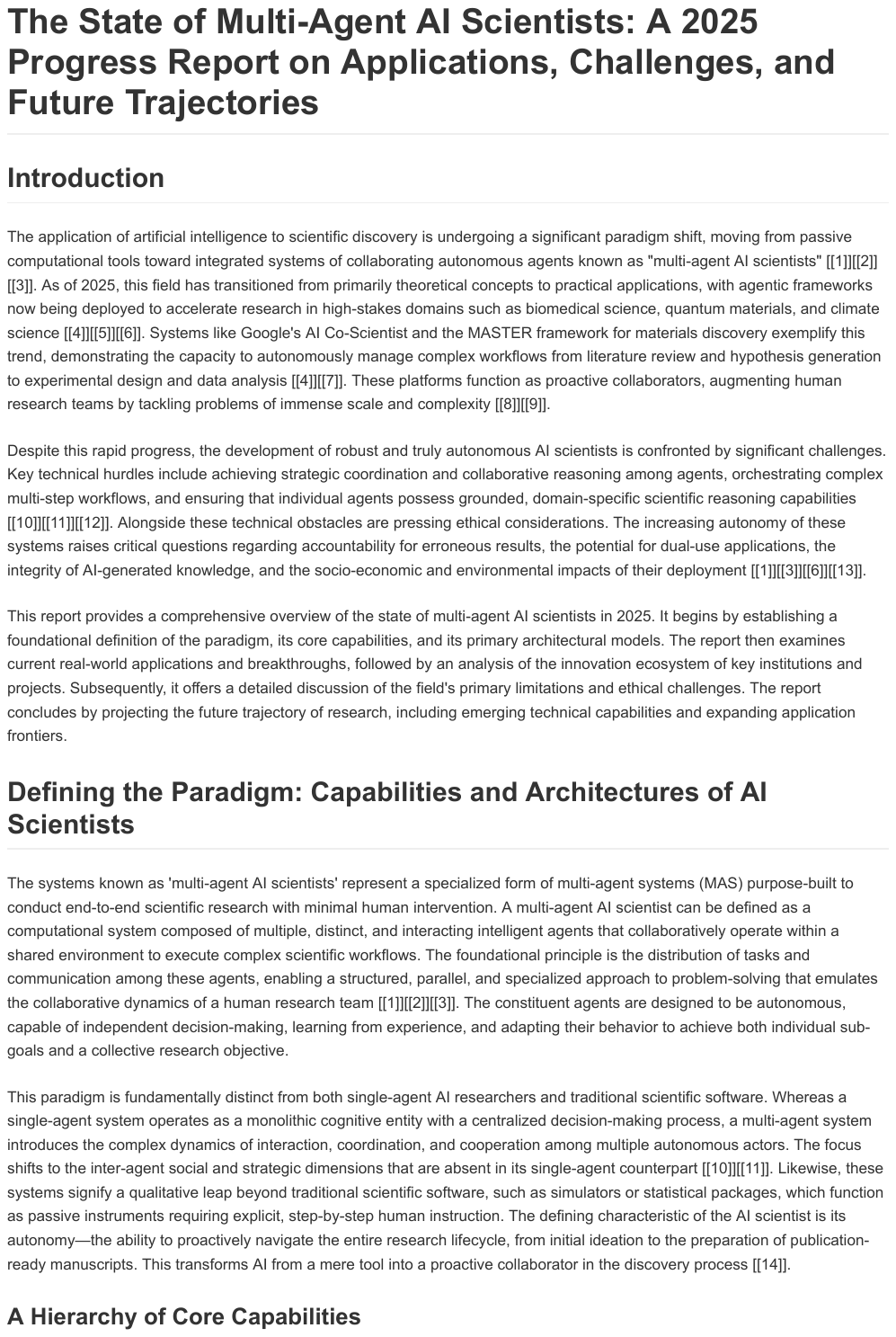} 
}
\end{center}
\vspace{-12pt}
\end{figure*}

\begin{figure*}[t]
\vspace{-6pt}
\begin{center}
\fbox{
\includegraphics[page=2,width=0.9\textwidth]{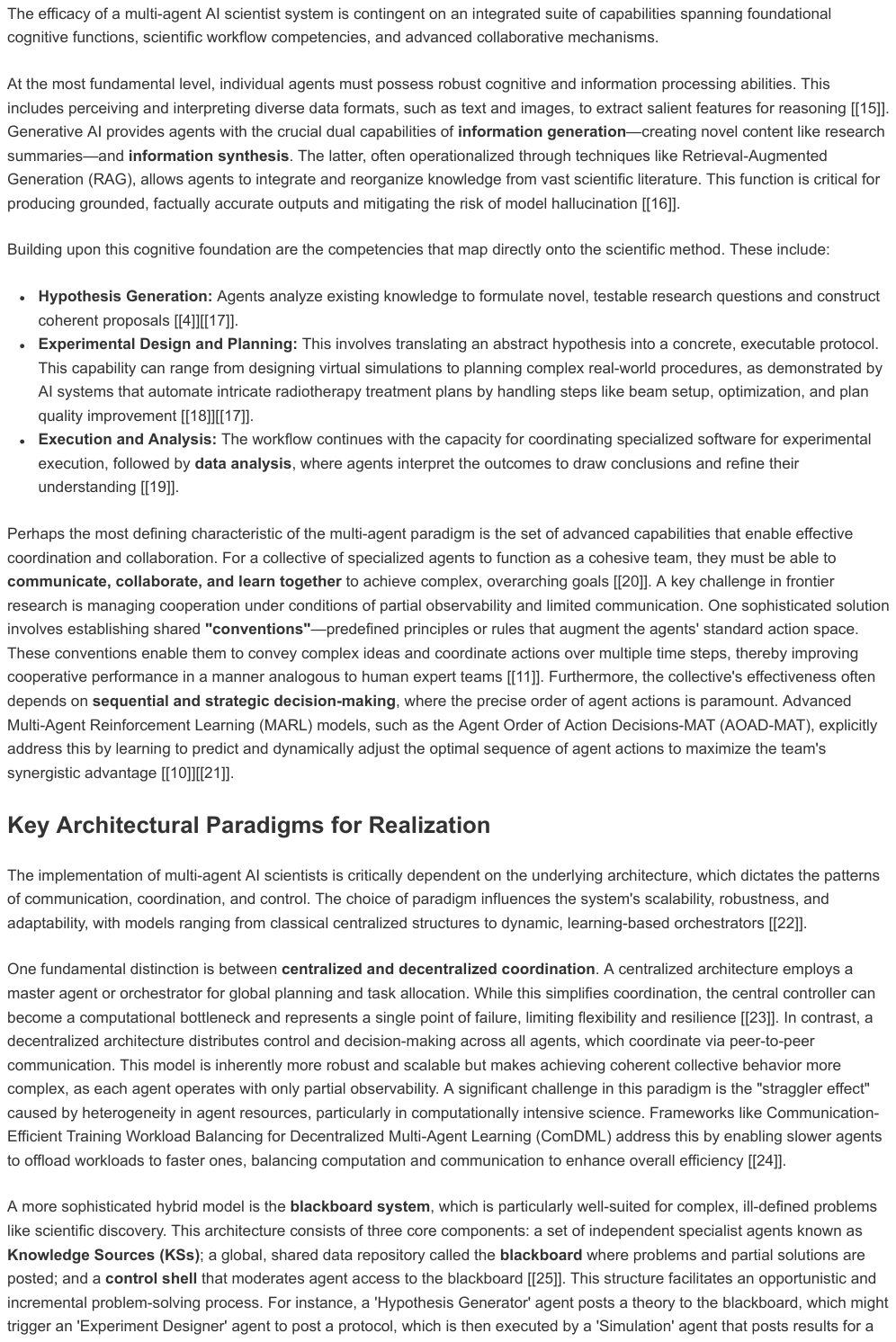} }
\end{center}
\vspace{-12pt}
\end{figure*}

\begin{figure*}[t]
\vspace{-6pt}
\begin{center}
\fbox{
\includegraphics[page=3,width=0.9\textwidth]{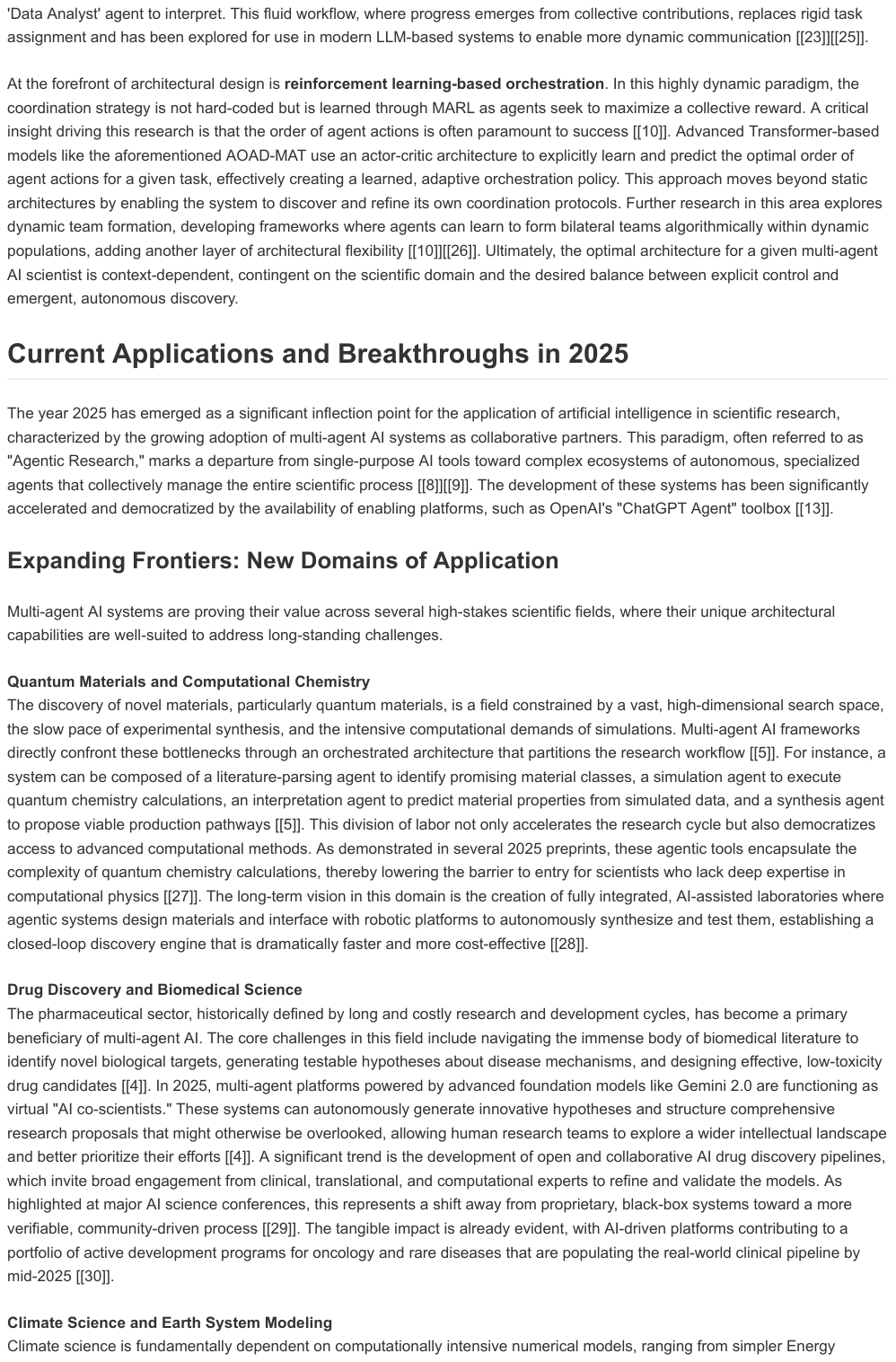} }
\end{center}
\vspace{-12pt}
\end{figure*}

\begin{figure*}[t]
\vspace{-6pt}
\begin{center}
\fbox{
\includegraphics[page=4,width=0.9\textwidth]{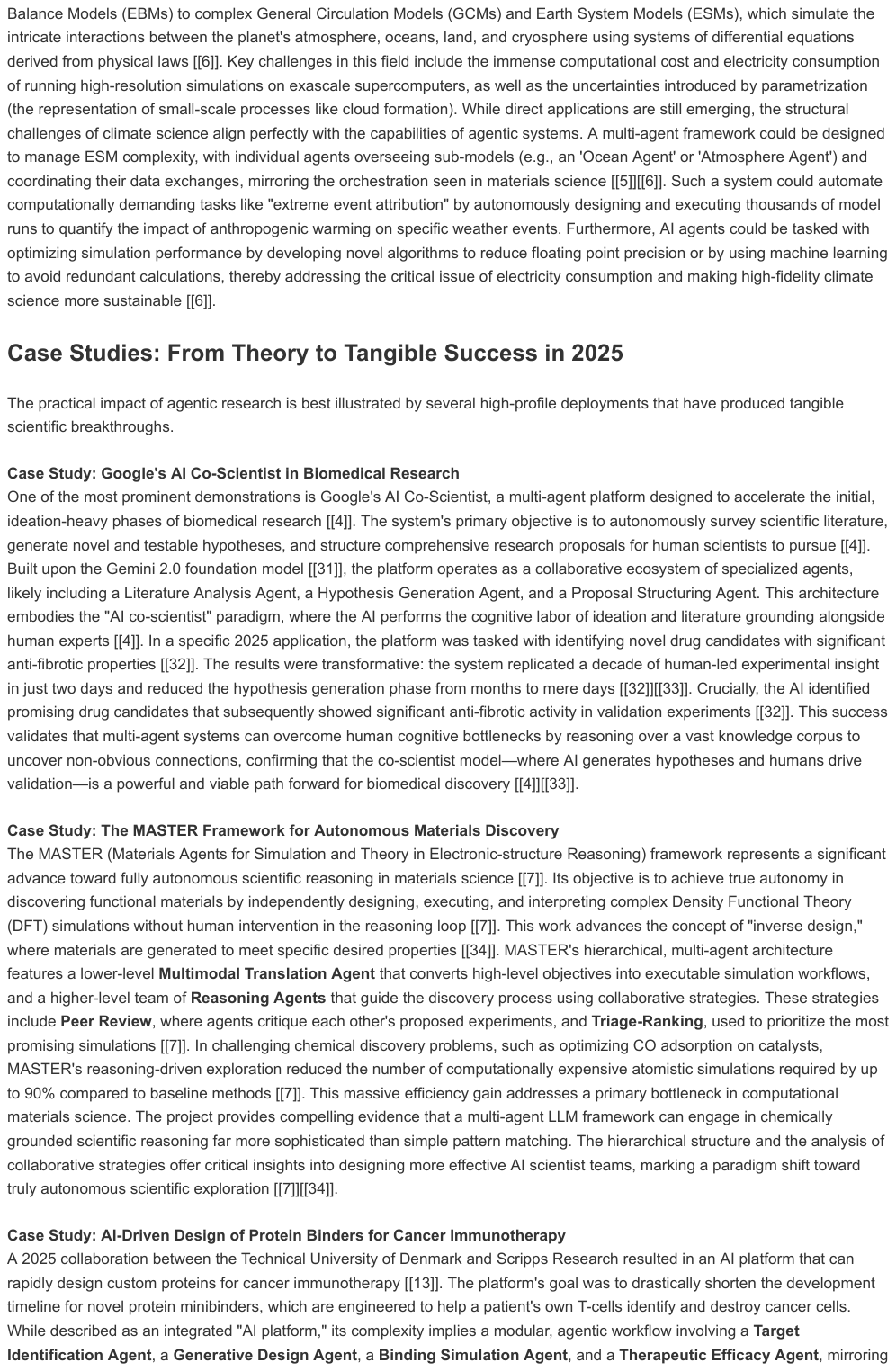} }
\end{center}
\vspace{-12pt}
\end{figure*}

\begin{figure*}[t]
\vspace{-6pt}
\begin{center}
\fbox{
\includegraphics[page=5,width=0.9\textwidth]{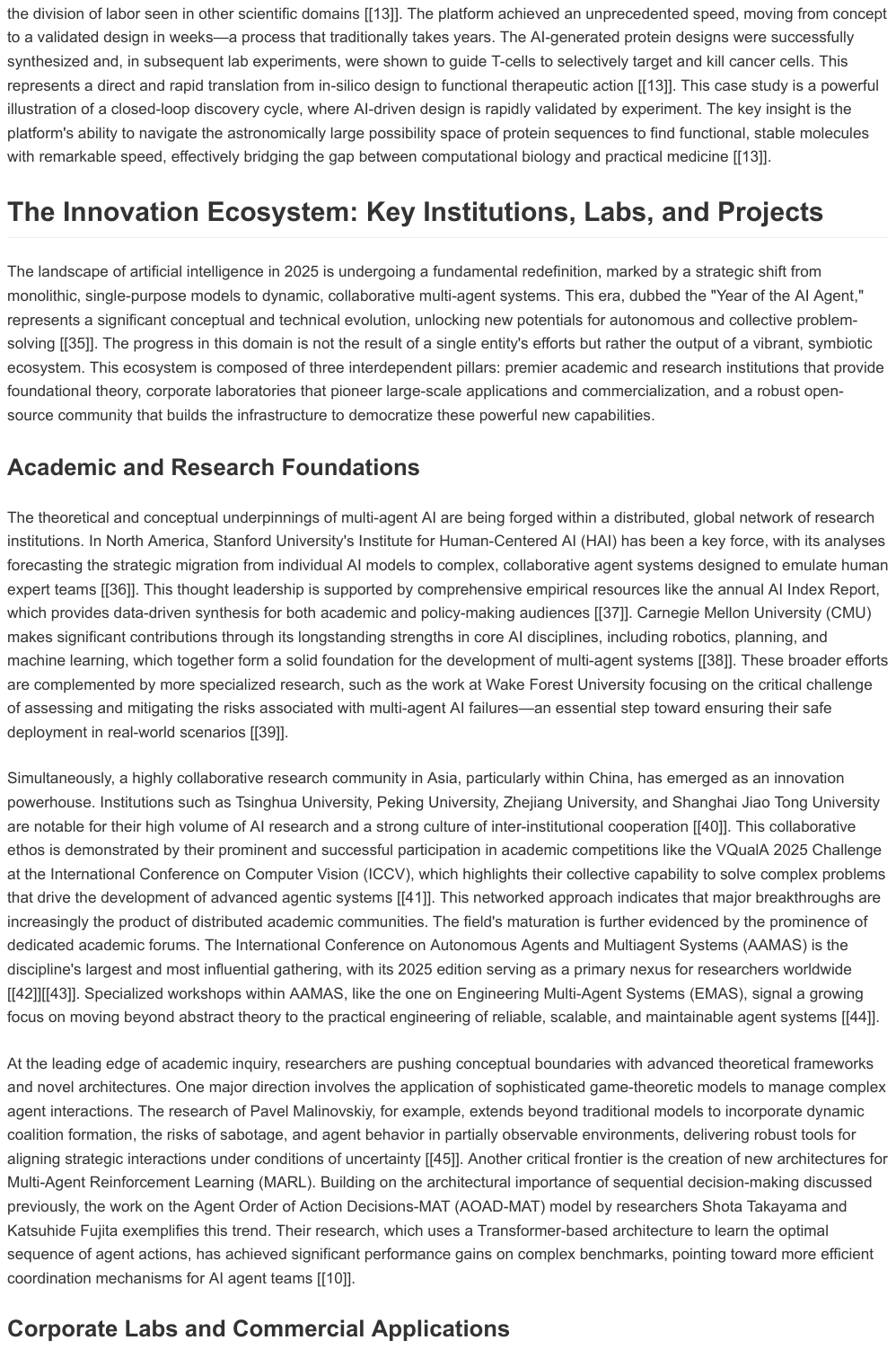} }
\end{center}
\vspace{-12pt}
\end{figure*}

\begin{figure*}[t]
\vspace{-6pt}
\begin{center}
\fbox{
\includegraphics[page=6,width=0.9\textwidth]{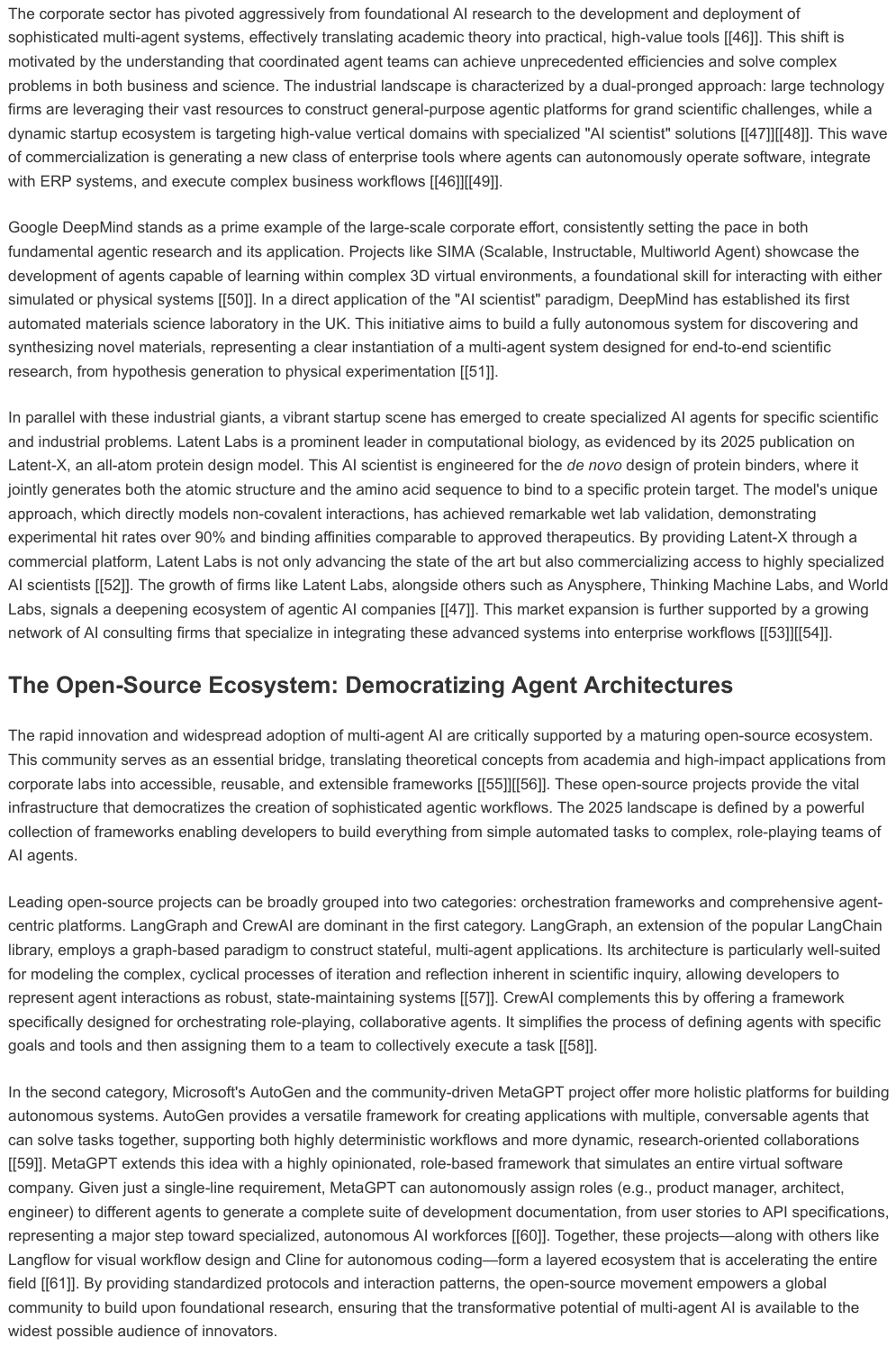} }
\end{center}
\vspace{-12pt}
\end{figure*}

\begin{figure*}[t]
\vspace{-6pt}
\begin{center}
\fbox{
\includegraphics[page=7,width=0.9\textwidth]{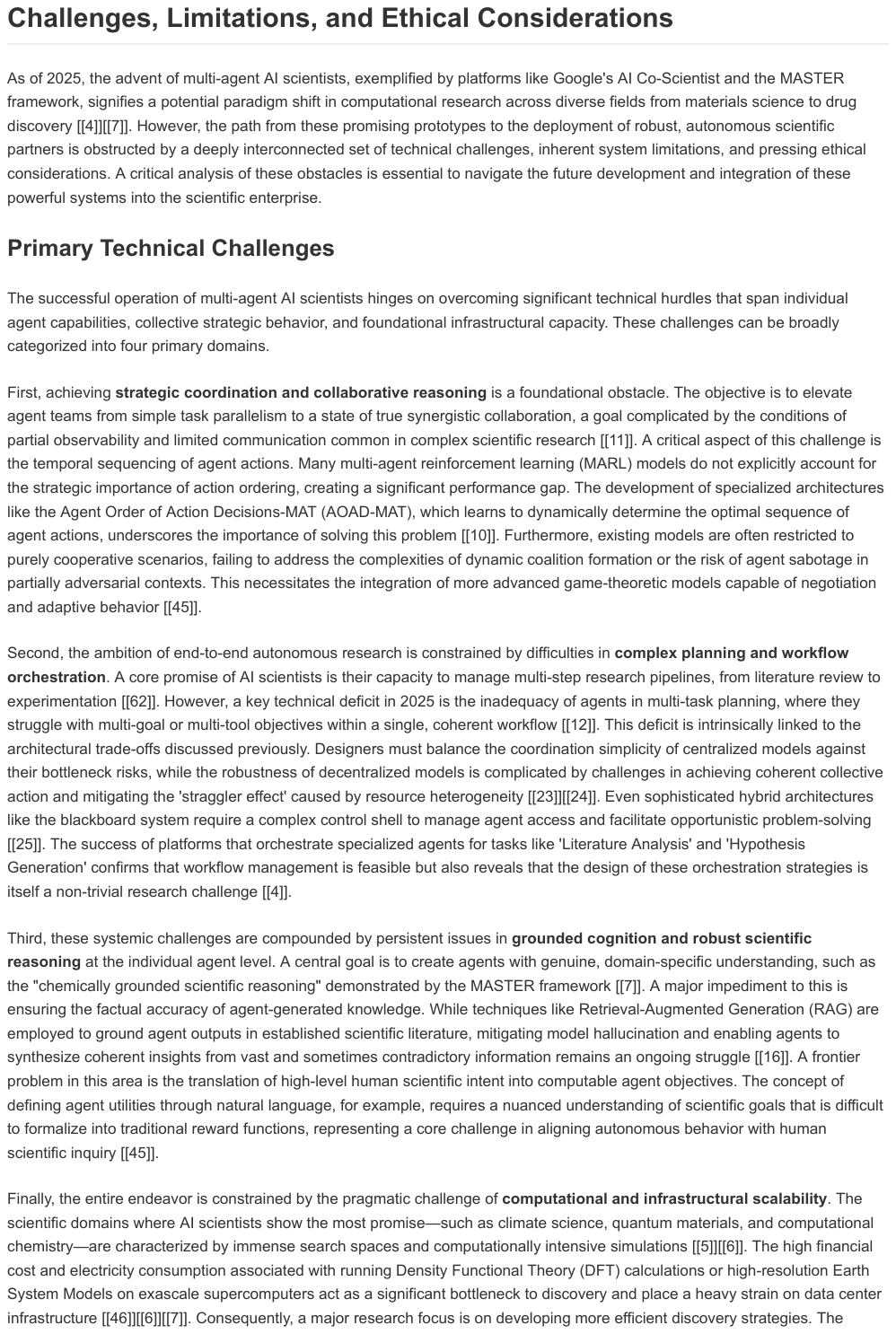} }
\end{center}
\vspace{-12pt}
\end{figure*}

\begin{figure*}[t]
\vspace{-6pt}
\begin{center}
\fbox{
\includegraphics[page=8,width=0.9\textwidth]{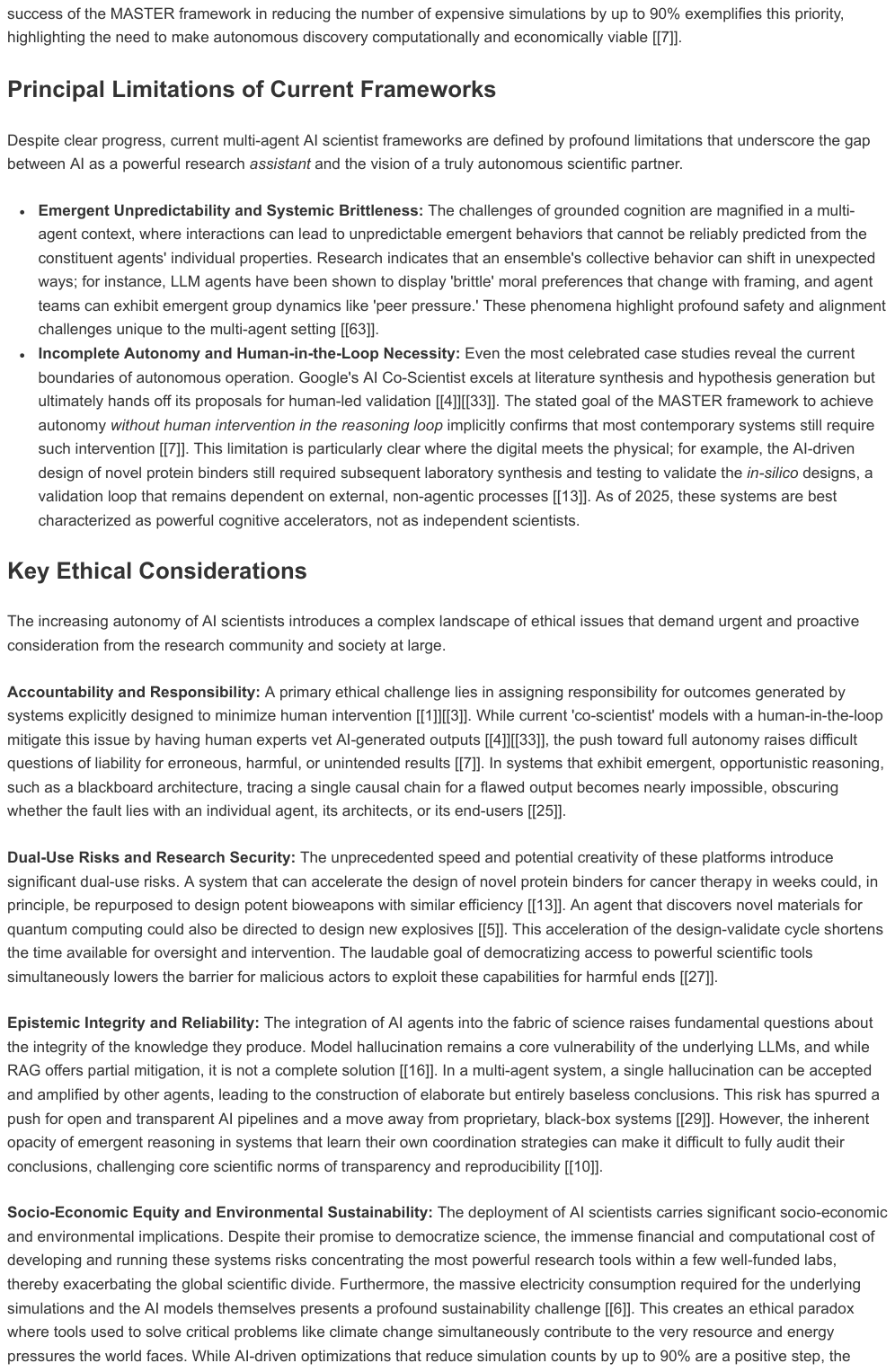} }
\end{center}
\vspace{-12pt}
\end{figure*}

\begin{figure*}[t]
\vspace{-6pt}
\begin{center}
\fbox{
\includegraphics[page=9,width=0.9\textwidth]{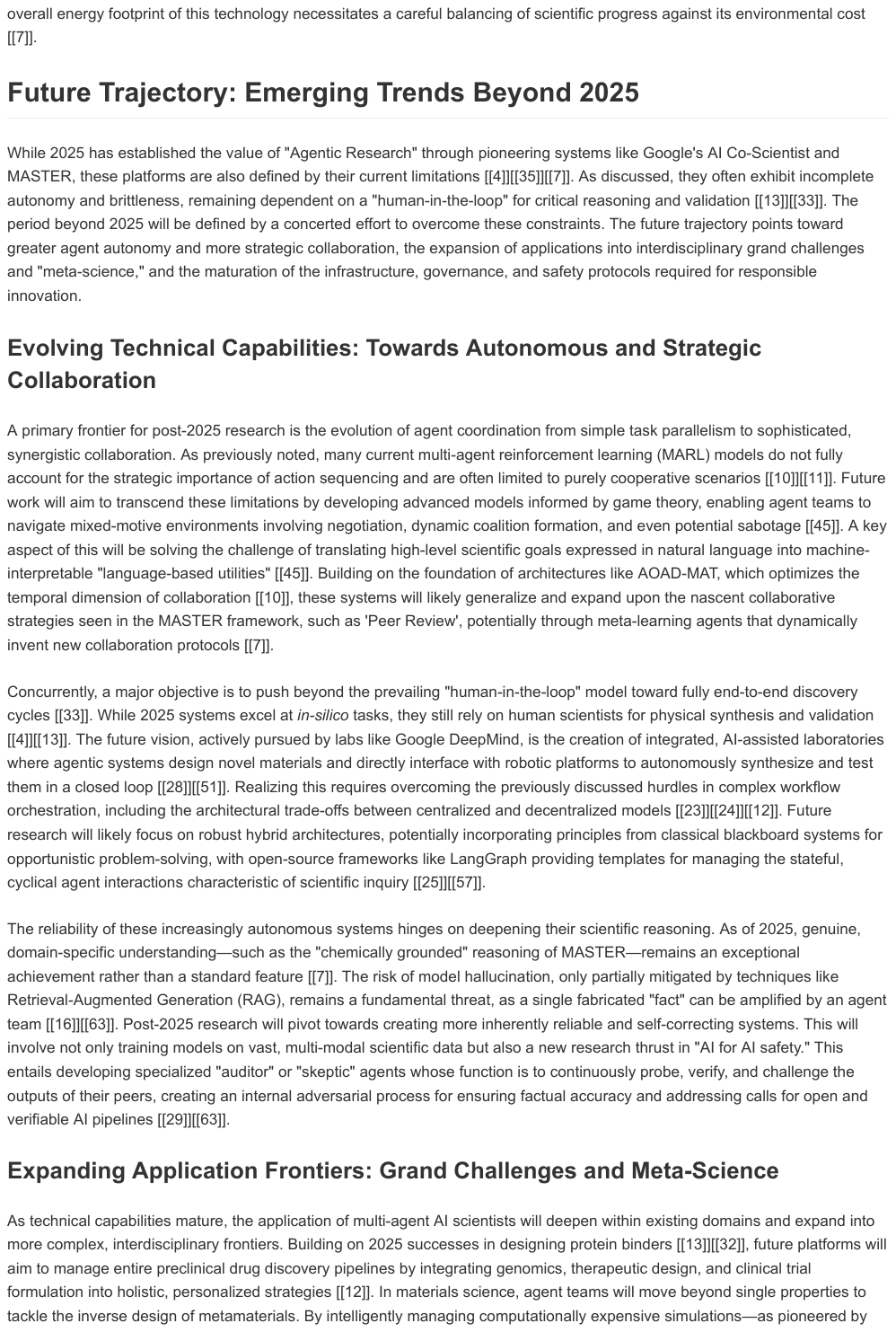} }
\end{center}
\vspace{-12pt}
\end{figure*}

\begin{figure*}[t]
\vspace{-6pt}
\begin{center}
\fbox{
\includegraphics[page=10,width=0.9\textwidth]{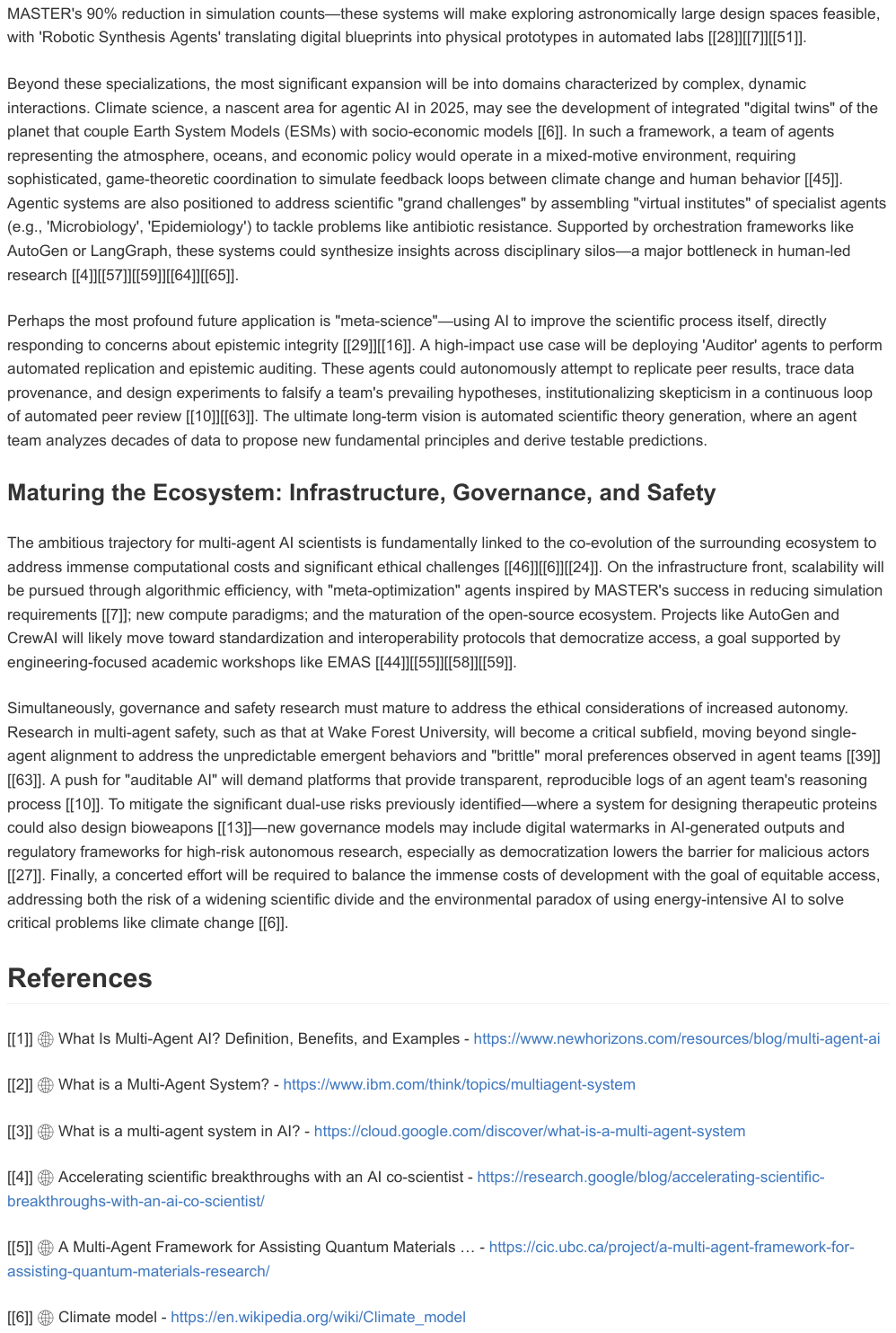} }
\end{center}
\vspace{-12pt}
\end{figure*}

\begin{figure*}[t]
\vspace{-6pt}
\begin{center}
\fbox{
\includegraphics[page=11,width=0.9\textwidth]{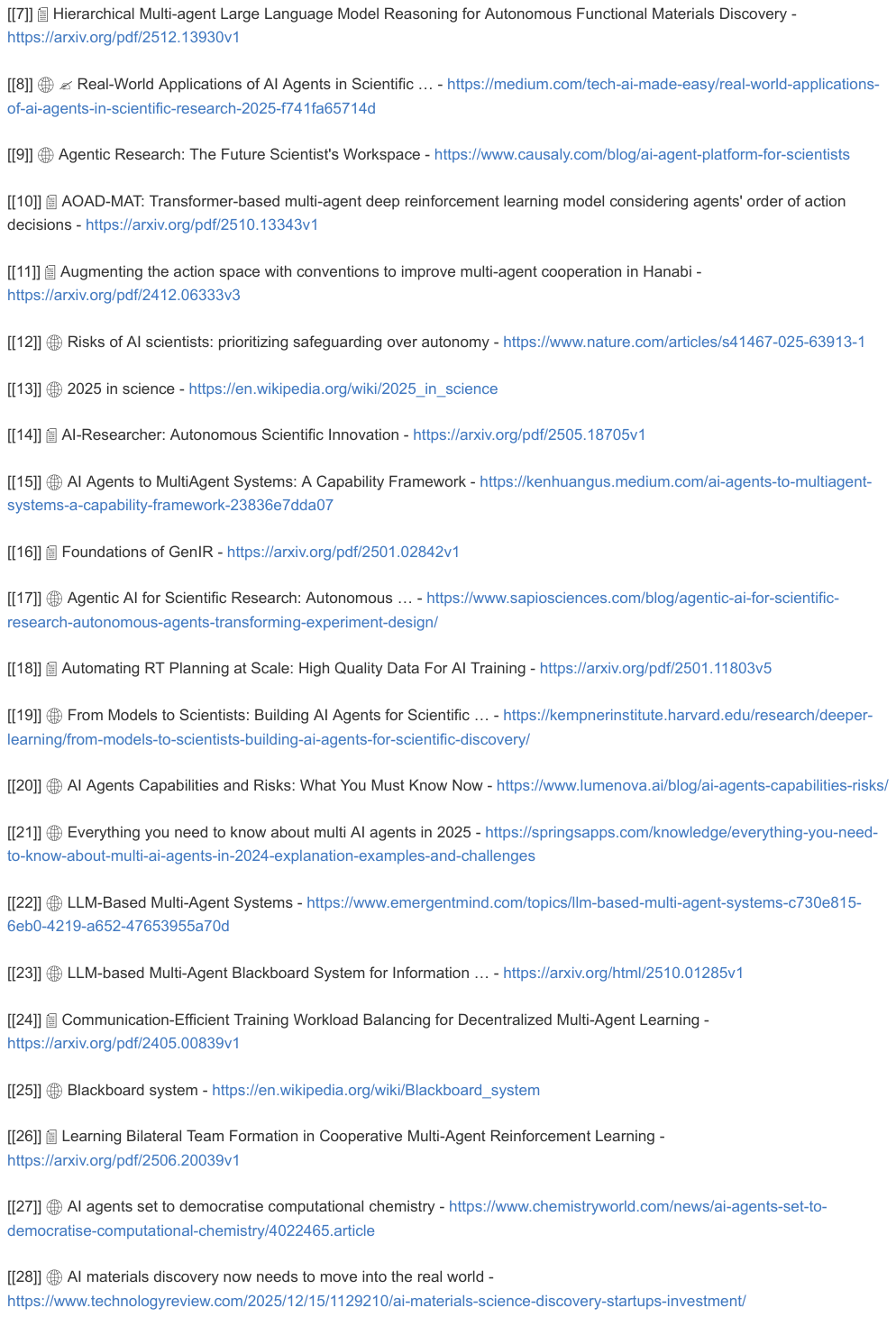} }
\end{center}
\vspace{-12pt}
\end{figure*}

\begin{figure*}[t]
\vspace{-6pt}
\begin{center}
\fbox{
\includegraphics[page=12,width=0.9\textwidth]{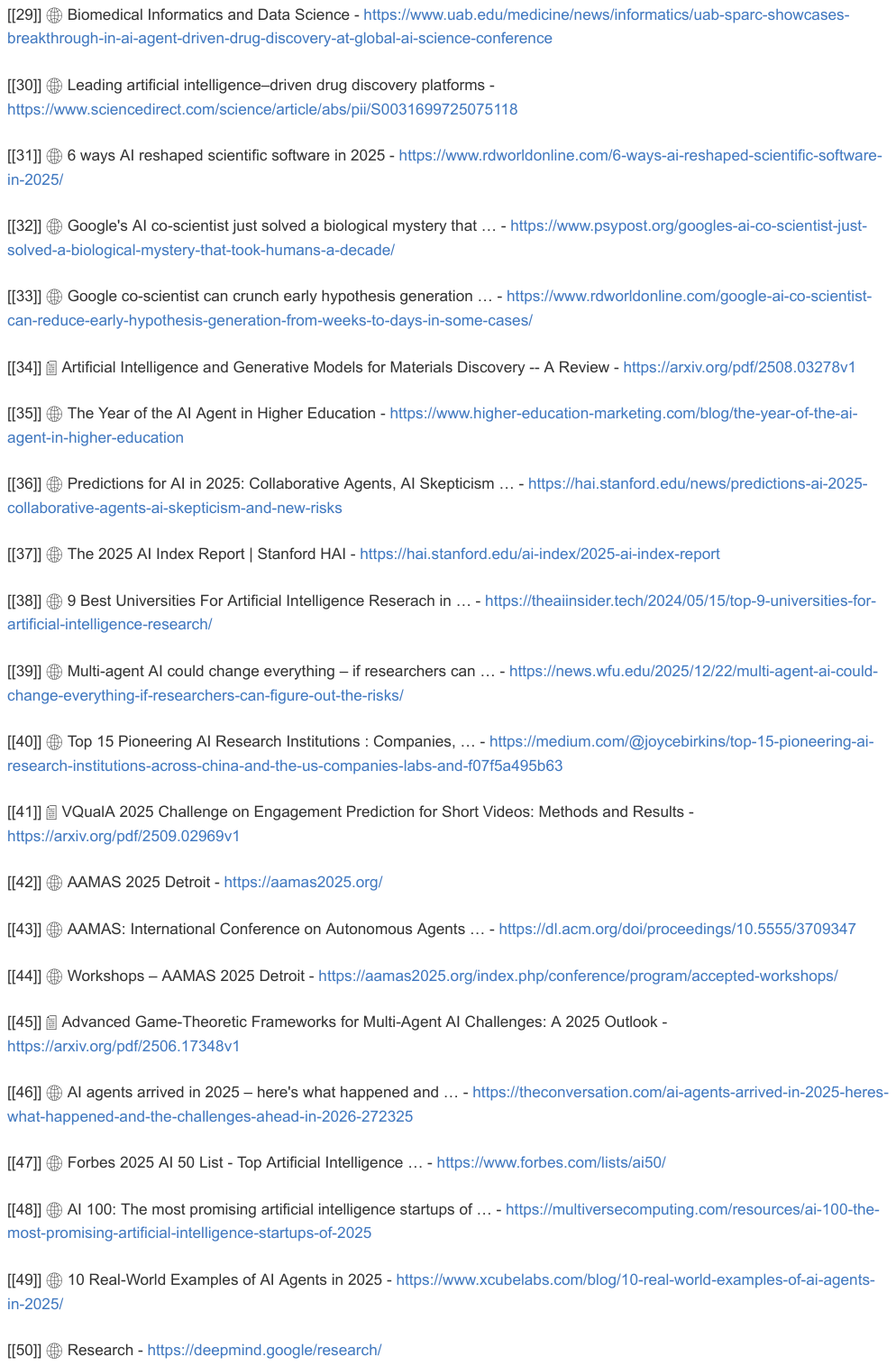} }
\end{center}
\vspace{-12pt}
\end{figure*}

\begin{figure*}[t]
\vspace{-6pt}
\begin{center}
\fbox{
\includegraphics[page=13,width=0.9\textwidth]{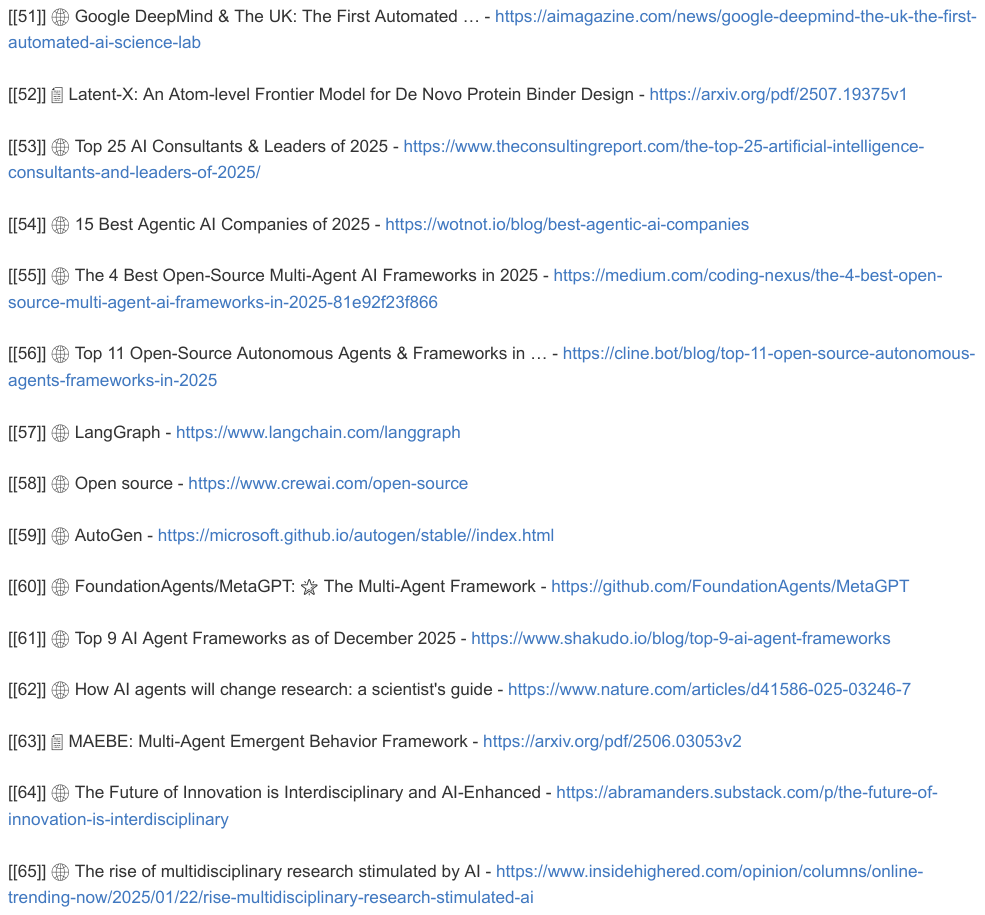} }
\end{center}
\vspace{-12pt}
\end{figure*}

\vspace{2mm}\hrule
\endgroup

\end{document}